\documentclass[%
 aip,
 amsmath,amssymb,
preprint,%
]{revtex4-1}

\usepackage{graphicx}
\usepackage{verbatim}
\usepackage{dcolumn}
\usepackage{bm}
\usepackage[utf8]{inputenc}
\usepackage[T1]{fontenc}
\usepackage{mathptmx}
\usepackage{etoolbox}
\usepackage{amsfonts}
\usepackage{multirow}
\usepackage{color}
\usepackage{amsthm}
\usepackage{algpseudocode}
\usepackage{float}
\usepackage[linesnumbered,ruled,vlined]{algorithm2e}
\newtheorem{thm}{Theorem}[section]

\newtheorem{prop}[thm]{Proposition}
\newtheorem{rem}[thm]{Remark}

\makeatletter
\def\@email#1#2{%
 \endgroup
 \patchcmd{\titleblock@produce}
  {\frontmatter@RRAPformat}
  {\frontmatter@RRAPformat{\produce@RRAP{*#1\href{mailto:#2}{#2}}}\frontmatter@RRAPformat}
  {}{}
}%
\makeatother
\begin{document}

\preprint{AIP/123-QED}

\title[Sample title]{MEP-Net: Generating Solutions to Scientific Problems with Limited Knowledge by Maximum Entropy Principle}

\author{Wuyue Yang}
\thanks{These authors have contributed equally to this work.}
\affiliation{Beijing Institute of Mathematical Sciences and Applications, Beijing, 101408, China}

\author{Liangrong Peng}
\thanks{These authors have contributed equally to this work.}
\affiliation{College of Mathematics and Data Science, Minjiang University, Fuzhou 350108, China}

\author{Guojie Li}
\affiliation{School of Mathematics, Sun Yat-sen University, Guangzhou 510275, China}          

\author{Liu Hong}
\thanks{Corresponding author: hongliu@sysu.edu.cn}
\affiliation{School of Mathematics, Sun Yat-sen University, Guangzhou 510275, China}


\begin{abstract}
Maximum entropy principle (MEP) offers an effective and unbiased approach to inferring unknown probability distributions when faced with incomplete information, while neural networks provide the flexibility to learn complex distributions from data. This paper proposes a novel neural network architecture, the MEP-Net, which combines the MEP with neural networks to generate probability distributions from moment constraints. We also provide a comprehensive overview of the fundamentals of the maximum entropy principle, its mathematical formulations, and a rigorous justification for its applicability for non-equilibrium systems based on the large deviations principle. Through fruitful numerical experiments, we demonstrate that the MEP-Net can be particularly useful in modeling the evolution of probability distributions in biochemical reaction networks and in generating complex distributions from data.
\end{abstract}


\maketitle

\begin{quotation}
Reconstruction of probability distributions from limited information is a fundamental challenge in various scientific fields, from statistical physics to machine learning. Traditional approaches often struggle with the trade-off between accuracy and data requirements. In this paper, we introduce the MEP-Net, a novel framework that synergistically combines the Maximum Entropy Principle (MEP) with neural networks to address this challenge. Using the theoretical foundations of the MEP and the flexible function approximation capabilities of neural networks, the MEP-Net demonstrates remarkable ability to generate complex probability distributions from moment constraints. Our approach not only bridges the gap between classical statistical methods and modern machine learning techniques, but also opens new avenues for solving inverse problems and integrating with other learning paradigms. Through a series of numerical experiments, we showcase the effectiveness of the MEP-Net in diverse scenarios, including high-dimensional distributions and time-dependent systems, highlighting its potential to enhance probabilistic modeling in data-limited regimes.
\end{quotation}

\section{Introduction}

Generative models aim to capture the statistical patterns in the training data by appropriately estimating and modeling the underlying probability distribution. They can generate new data samples based on the learned probability distribution, enabling various applications such as data generation, anomaly detection, and probabilistic inference. With the rise of big data and deep learning, various neural network-based generative models, such as the Generative Adversarial Networks (GANs) \cite{goodfellow2020generative}, Variational Auto Encoders (VAEs)\cite{kingma2013auto}, Transformers\cite{vaswani2017attention}, Diffusion models\cite{ho2020denoising, song2020score}, have emerged as powerful tools capable of learning complex distributions or generating new data samples.

Many scientific problems, ranging from statistical physics to molecular dynamics and from biological systems to climate patterns, often involve scenarios where only partial information is available, typically in the form of statistical moments or other constraints. Traditional statistical methods may fail to accurately capture the true distributions. Moreover, modern machine learning techniques often generate physically incredible results when applied to scientific problems with limited data.


Classically, there is a notable theoretical framework rooted in statistical physics for inferring the desired probability distribution of a system with insufficient information, which is known as the Maximum Entropy Principle (MEP). From a historical point of view, three cornerstones for the development of MEP should be mentioned: Boltzmann's maximum-multiplicity theory and Gibb's ensemble theory, Jaynes's formulation \cite{jaynes1957information,jaynes1957informationII}, as well as Shore and Johnson's viewpoint\cite{}. As pointed out by Beck \cite{beck2009generalised}:``\textit{The entire formalism of statistical mechanics can be regarded as being based on maximizing the entropy of the system under consideration subject to suitable constraints}.'' Far beyond its original scope for predicating  equilibrium states of systems, the MEP has already demonstrated its power in multiple disciplines, including the earth and climate system \cite{kleidon2004non}, ecology \cite{harte2011maximum}, reliability science \cite{rocchi2017reliability}, and more recently image reconstruction \cite{prakash2019maximum}, time series analysis \cite{marchenko2018towards}, model inference from data \cite{gresele2017maximum}, and deep reinforcement learning \cite{haarnoja2018soft}. Interested readers are referred to the review \cite{Presse2013Principles} for more details. 

We notice that no matter in the literature or in practice, there is no generally accepted deep learning framework to implement the Maximum Entropy Principle till now. A key bottle we suspect is the lack of a suitable entropy function. In addition, the concrete form of constraints also matters a lot. To overcome these difficulties, in this paper we propose a practical way to construct an effective entropy loss function based on the concept of fixed-point iteration. This entropy loss function takes the form of Kullback-Leibler divergence, which therefore is always non-negative and reaches its minimum if and only if the neural networks converge. We further test the reconstruction ability of different moments (or data constraints), and the binomial functions show a much better performance than polynomials and other functions. In this way, we successfully combine the Maximum Entropy Principle with neural networks. The MEP offers a rigorous and unbiased approach to infer the desired probability distributions with respect to incomplete information, while neural networks provide the flexibility to learn complex distributions from data samples. 

The rest of the paper is structured as follows. In Section \ref{Fundamentals of Maximum Entropy Principle}, we provide the theoretical background of the Maximum Entropy Principle and its applications in reconstruction of various classci probability distributions. 
In Section \ref{The Architecture of MEP-Net}, we illustrate the architecture of our MEP-Net including the screening of feature libraries and loss functions. In Section \ref{Numerical Experiments}, we present a series of numerical examples about multi-modal statistical distributions, chemical master equations, particle diffusion in a confined domain, and the Allen-Cahnn model for binodal decomposition. Section \ref{Conclusion and Discussion} is a brief Conclusion and Discussion. 

\section{Fundamentals of Maximum Entropy Principle}
\label{Fundamentals of Maximum Entropy Principle}

\subsection{Mathematical Formulation}
\emph{Entropy}. Assume that the state space of the system is continuous (the discrete case can be treated analogously). Entropy is a measure of the amount of information or uncertainty in a system, which can be represented as a functional of a probability density function (pdf) $p$, i.e. $\mathcal{F}[p].$ The Shannon entropy, denoted as $\mathcal{S}[p]$, is defined as, 
\begin{equation}
\mathcal{S}[p]=-\mathbb{E}_p[\ln p(\vec{x})],
\end{equation}
where $\mathbb{E}_p[\cdot]$ denotes the expectation with respect to the probability $p$, and $\vec{x}\in \mathbb{R}^d$. For a continuous pdf, it is written explicitly as 
\begin{equation*}
\mathcal{S}[p]=-\int_{\mathbb{R}^d} p(\vec{x}) \ln p(\vec{x}) d\vec{x}. 
\end{equation*}
Besides the Shannon entropy, there are two other significant entropies in information theory. On the one hand, the relative entropy, also known as the Kullback-Leibler (KL) divergence\cite{}, provides a measure of the distance of a probability distribution $p(\vec{x})$ against another reference probability distribution $q(\vec{x})$,
\begin{equation}
\label{KL}
\mathcal{D}(p \| q)=\mathbb{E}_p\left[\log \frac{p(\vec{x})}{q(\vec{x})}\right].
\end{equation}
On the other hand, the cross-entropy measures the difference between two probability distributions,
\begin{equation*}
\mathcal{H}(p,q)=-\mathbb{E}_p[\log q(\vec{x})],
\end{equation*}
which satisfies the relation $\mathcal{H}(p,q)=\mathcal{S}[p]+\mathcal{D}(p \| q)$. 




\emph{General formulation.}
According to the MEP, the maximization of the Shannon entropy function $\mathcal{S}[p]$ under constraints reads \cite{Presse2013Principles}
\begin{equation}
\label{conditional maximization}
\begin{split}
    &\max_{p(\vec{x})} 
    \mathcal{S}[p]=-\int_{\mathbb{R}^d} d\vec{x}p(\vec{x}) \ln p(\vec{x}),\\ 
    &s.t.,\quad\int_{\mathbb{R}^d}d\vec{x} p(\vec{x})= 1,\\
    &\qquad\int_{\mathbb{R}^d} d\vec{x} p(\vec{x}) f_i(\vec{x})=\tilde{f}_i,~ i=1,\cdots,M,
\end{split}
\end{equation}
where the first constraint corresponds to the normalization condition of the pdf, the rest correspond to the observables $f_i$ from the data. For instance, the choice of $f_i(\vec{x})=x^i$ leads to the $i$-th order moment about origin. The above system \eqref{conditional maximization} turns into a unconstrained maximization problem by using the method of Lagrangian multiplier, 
\begin{eqnarray*}
&&\max_{p(\vec{x});\lambda_i}L[p;\lambda_i]\\
&&=-\int_{\mathbb{R}^d} d\vec{x} p(\vec{x}) \ln p(\vec{x}) + \lambda_0 \left( \int_{\mathbb{R}^d}p(\vec{x})d\vec{x}-1 \right)
+ \sum_{i=1}^M  \lambda_i\left( \int_{\mathbb{R}^d} d\vec{x} p(\vec{x}) f_i(\vec{x})-\tilde{f}_i
\right),
\end{eqnarray*}
where $L[p;\lambda_i]$ is the Lagrangian, and $\lambda_i$'s are the Lagrangian multipliers for the corresponding constraints. The variation (resp. partial differential) of $L$ with respect to $p(\vec{x})$ (resp. to  $\lambda_i$) $0=\frac{\delta L}{\delta p}=\frac{\partial L}{\partial \lambda_i},~(i=0,1,\cdots, M)$ yields an optimal pdf,
\begin{equation}
p^*(\vec{x})=\exp\left(\lambda_0 -1 + \sum_{i=1}^M \lambda_i f_i(\vec{x})\right),
\end{equation}
in which those Lagrangian multipliers are determined through the following algebraic relations 
\begin{align*}
    &e^{\lambda_0 - 1} \int_{\mathbb{R}^d}d\vec{x}  \left( e^{\sum_{i=1}^M \lambda_i f_i(\vec{x})}\right) =1,\\
    &e^{\lambda_0 - 1} \int_{\mathbb{R}^d}d\vec{x}  \left(f_i(\vec{x}) e^{\sum_{i=1}^M \lambda_i f_i(\vec{x})}\right) =\tilde{f}_i, ~(i=1,\cdots, M).
\end{align*}
By elimination of the parameter $\lambda_0$, this optimal pdf can be alternatively expressed as
\begin{equation}\label{MEP}
p^*(\vec{x})=\frac{\exp(\sum_{i=1}^M \lambda_i f_i(\vec{x}))}{\int_{\mathbb{R}^d}d\vec{x} \exp(\sum_{i=1}^M \lambda_i f_i(\vec{x}))},
\end{equation}
where the denominator ${\int_{\mathbb{R}^d}d\vec{x} \exp(\sum_{i=1}^M \lambda_i f_i(\vec{x}))}$ is interpreted as the partition function in statistical physics. 

\emph{Relation with minimum relative entropy principle.} 
Let $q(\vec{x})$ denote the prior probability density function. According to the definition in Eq. \eqref{KL}, the relative entropy between the prior pdf $q(\vec{x})$ and the posterior pdf $p(\vec{x})$ is $\mathcal{D}(p \| q)=\int_{\mathbb{R}^d} d\vec{x} p(\vec{x}) \ln[p(\vec{x})/q(\vec{x})]$. The minimization of the relative entropy $\mathcal{D}(p \| q)$ under the same constraints as those in MEP yields an optimal distribution 
\begin{equation}\label{Minrela}
    \tilde{p}^*(\vec{x})
    =q(\vec{x})\exp\left(-\lambda_0 +1 - \sum_{i=1}^M \lambda_i f_i(\vec{x})\right)
    =\frac{q(\vec{x})\exp( -\sum_{i=1}^M \lambda_i f_i(x))}{\int_{\mathbb{R}^d}d\vec{x} q(\vec{x})\exp( -\sum_{i=1}^M \lambda_i f_i(\vec{x}))},
\end{equation}
where the parameter $\lambda_0$ is eliminated in the last equality. The optimal distributions in Eqs. \eqref{MEP} and \eqref{Minrela} are intimately related, since the multiplier $q(\vec{x})$ serves as a weight function (or called degeneracy in statistical physics), which is set to be uniformly distributed in the MEP. 

\subsection{Formal Derivation Based on Large Deviations Principle}
When restricted to equilibrium systems, the validity of the MEP is guaranteed by the second law of thermodynamics. However, once extended to a non-equilibrium system, why and when the MEP can still be applied are questionable. Here we provide a more rigorous justification of the MEP with the help of the large deviations principle (LDP), which holds even for most non-equilibrium cases. 

Consider a sequence of independently and identically distributed (IID) random variables $\vec{X}=(X_1,X_2,\cdots,X_n)$. For simplicity, we further suppose that they take discrete values within a finite set, $\Gamma=\{1,2,\cdots,m\}$. So that we have $P(X_i=j)=q_j>0$, where $i\in\{1,2,\cdots,n\},~j\in\Gamma$. Now the problem addressed by the MEP becomes how to find out the common probability function $q_j$ based on the knowledge of observations on random variables $\vec{X}$ (e.g. the moments).

Clearly, there is no unique solution if only the information about finite observations is taken into consideration. So the MEP introduces the entropy function as an additional criterion for selection. Alternatively, according to laws of large numbers, it is well-known that the frequency of $n$ IID random variables approaches to their probability as $n\rightarrow\infty$. Therefore, it is expected an unique optimal distribution can be determined during the procedure of infinitely repeated observations. The analysis of corresponding asymptotic limit gives the desired entropy function (relative entropy, to be exact) in the MEP. In this way, the statement of MEP becomes a natural consequence of the LDP.

To be concrete, we introduce the following counting frequency
\begin{eqnarray*}
K_{n,j}(\vec{X})=\frac{1}{n}\sum_{i=1}^{n}\delta_{X_i,j},\quad j\in\Gamma.
\end{eqnarray*} 
Then it can be proved that as $n\rightarrow\infty$, $K_{n,j}$ obeys the large deviations principle, which guarantees the existence of the following large deviations rate function (LDRF)\cite{} 
\begin{eqnarray*}
I(\vec{l})=\lim_{n\rightarrow\infty}-\frac{1}{n}P(\vec{K}_{n}(\vec{X})=\vec{l}),
\end{eqnarray*} 
where $\vec{K}_{n}=({K}_{n,1}, {K}_{n,2}, \cdots, {K}_{n,m})$  forms a random vector, and $\vec{l}\in \mathbb{R}^m$. 
According to Sanov's theorem, the LDRF can be explicitly written out, \textit{i.e.}
\begin{eqnarray}
I(\vec{l})=\sum_{j\in\Gamma}l_j\ln\left(\frac{l_j}{q_j}\right).
\end{eqnarray} 
It is easily seen that $I(\vec{l})$ is a non-negative, convex function, which reaches its minimum if and only if $l_j=q_j$ for each $j\in\Gamma$\cite{}. Therefore, the LDRF $I(\vec{l})$ turns out to be the relative entropy $\mathcal{D}(\vec{l} \| \vec{q})$ in the MEP. And the estimated pdf obtains its optimal solution $\vec{q}=(q_1, q_2, \cdots, q_m)$ once the entropy function reaches its extreme values.  It should be noted that Sanov's theorem holds even when $\Gamma$ is a set with infinite discrete or even continuous values\cite{}. Therefore, the above justification for MEP based on LDP can be easily extended to these more complicated cases, which will not be addressed here.

Our argument till now is only valid for non-constrained cases. To take the influence of constraints obtained from the observations into consideration, we refer to the famous contraction principle\cite{}. Given a group of constraints on the solution $\vec{l}$, i.e. $\sum_{j\in\Gamma}l_jf_i(j)=\tilde{f}_i$ where $f_i$ is a real function (like the expectation and variance) and $i=1,2,\cdots,M$. Then the new LDRF for probabilities fulfilling those constraints is given by 
\begin{eqnarray*}
I^{con}(\vec{l})=\inf_{\vec{l}:\sum_{j\in\Gamma}l_jf_i(j)=\tilde{f}_i, \forall i\in\{1,2,\cdots,M\}} I(\vec{l}).
\end{eqnarray*} 
The above optimization problem can be solved by using the method of Lagrangian multipliers, just as we do in the MEP. Therefore, as long as $n\rightarrow\infty$, we can build the whole MEP on the mathematical foundation of LDP.

\subsection{Optimal PDFs Corresponding to Various Entropy Functions and Constraints}
To find out the optimal probability distribution that fits to the data in the best way, the
entropy function and the constraints must be carefully designed.
Here are some general suggestions on this point based on the intrinsic physical properties of the system. Firstly, we need to determine whether the system is extensive or non-extensive, which mainly depends on the nature of interactions. In the presence of either long-range interactions or memory effects, the system will be non-extensive. The classical Boltzmann-Gibbs (BG) entropy is usually adopted for extensive systems, while the non-extensive entropy, such as the Tsallis \cite{tsallis1988possible} or Renyi entropy, is more appropriate for the non-extensive ones. Secondly, the choice of constraints is closely related to the observable variables of a system. In general, moments, such as the mean and variance, are adopted as observable variables.   

To clarify the intrinsic correspondence between the optimal probability distribution function and entropy-constraint pairs in the MEP, we make a comprehensive summary on the results of MEP in Tables \ref{Tab1} and \ref{Tab2}. Meanwhile, the details of derivation are left in Appendices  \ref{Rigorous derivations of various continuous distribution} and \ref{Rigorous derivations of various discrete distribution}. Although some optimal probability distributions have been addressed in previous works, such as in the Refs. \cite{park2009maximum, kagan1973characterization}, here we provide Tables I and II in a more rigorous and comprehensive fashion. 

Generally speaking, the optimal probability distributions listed in Tables \ref{Tab1} and \ref{Tab2} can be divided into two different groups. The first group includes the normal, exponential distributions, etc., in which the number of free parameters is equal to that of the constraints other than the probability normality. Take the 1-dimensional normal distribution for instance, this number is $2$, consisting of constraints on the expectation ($\mathbb{E}[X]$) and variance ($\mathbb{E}[X^2]-(\mathbb{E}[X])^2$). The other group includes the Gamma, Rayleigh distributions, etc., in which the number of free parameters is less than that of the constraints other than the probability normality. For the Rayleigh distribution, the only free parameter is its scale parameter, while there are two constraints on both the squared expectation ($\mathbb{E}[X^2]$) and the natural logarithmic expectation ($\mathbb{E}[\ln X]$). This facts suggests that a trade-off between these two constraints may be necessary. 

\begin{table}[]
\setlength\extrarowheight{10pt}
\setlength{\tabcolsep}{2pt}
\scriptsize
\centering
\caption{Continuous optimal pdfs corresponding to various entropy functions and constraints}
\label{Tab1}
\begin{tabular}{llllllll}
\hline
Distribution& Entropy& Constraints& Optimal Distribution\\
\hline
Uniform&
$S[a,b]$&
$\mathbb{E}1$&
$\frac{1}{b-a}\mathbb{I}_{a\leq x \leq b}$\\
\hline
Exponential&$S[0,\infty)$
&$\mathbb{E}1, \mathbb{E}X$
&$\frac{1}{\mathbb{E}X}\exp\left(-\frac{x}{\mathbb{E}X}\right)\mathbb{I}_{x\geq0}$\\
\hline
Normal&
$S(-\infty,\infty)$&
$\mathbb{E}1, \mathbb{E}X, \mathbb{E}X^2$
&$\frac{1}{\sqrt{2\pi(\mathbb{E}X^2 - (\mathbb{E}X)^2)}}\exp\left({-\frac{(x -\mathbb{E}X)^2}{2(\mathbb{E}X^2 - (\mathbb{E}X)^2)}}\right)$\\
\hline
Higher-order&
$S(-\infty,\infty)$
&$\mathbb{E}1, \mathbb{E}X, \cdots, \mathbb{E}X^M$
&$\exp(\lambda_0 - 1)\exp\left(\sum_{i=1}^{M} \lambda_ix^i\right)$\\
\hline
Gamma&
$S(0,\infty)$&
$\mathbb{E}1, \mathbb{E}X, \mathbb{E}\ln X$&
$\frac{(-\lambda_1)^{\lambda_2+1}}{\Gamma(\lambda_2+1)}x^{\lambda_2}e^{\lambda_1 x} \mathbb{I}_{x>0}$\\
\hline
Erlang&
$S[0,\infty)$&
$\mathbb{E}1, \mathbb{E}X, \mathbb{E}\ln X$&
$\frac{(-\lambda_1)^{\lambda_2+1}}{\Gamma(\lambda_2+1)}x^{\lambda_2}e^{\lambda_1 x} \mathbb{I}_{x\geq 0}~(\lambda_2 \in \mathbb{Z}_{\geq 0})$\\
\hline
Beta&$S[0,1]$
&$\mathbb{E}1, \mathbb{E}\ln X, \mathbb{E}\ln(1-X)$
&$\frac{\Gamma(\lambda_1+\lambda_2+2)}{\Gamma(\lambda_1+1)\Gamma(\lambda_2+1)}x^{\lambda_1}(1-x)^{\lambda_2}\mathbb{I}_{0<x<1}$\\
\hline
Log-normal&
$S(0,\infty)$&
$\mathbb{E}1, \mathbb{E}\ln X, \mathbb{E}(\ln X)^2$
&$e^{\lambda_0-1}x^{\lambda_1} \exp\left[\lambda_2 (\ln x)^2\right]\mathbb{I}_{x>0}$\\
\hline
Double exp. (Laplace)&
$S(-\infty,\infty)$& 
$\mathbb{E}1, \mathbb{E}|X-\mu|$
&$\frac{1}{2\langle|X-\mu| \rangle }\exp{\left(-\frac{\langle|x-\mu| \rangle}{\langle|X-\mu| \rangle} \right)}$\\
\hline
$\chi^2$&
$S[0,\infty)$& 
$\mathbb{E}1, \mathbb{E}X, \mathbb{E}\ln X=\frac{\Gamma'({\langle X \rangle}/2)}{\Gamma({\langle X \rangle}/2)}+\ln 2$&
$\frac{1}{2^{{\langle X \rangle}/{2}} \Gamma({\langle X \rangle}/{2})}x^{{\langle X \rangle}/{2}-1}e^{-{\langle X \rangle}/{2}} \mathbb{I}_{x\geq 0}$\\
\hline
F&
$S[0,\infty)$&
$\mathbb{E}1, \mathbb{E}\ln X, \mathbb{E}\ln (1+\frac{d_1}{d_2}X)$&
$e^{\lambda_0 -1} x^{\lambda_1} (1+\frac{d_1}{d_2}x)^{\lambda_2} \mathbb{I}_{x\geq 0}$\\
\hline
Rayleigh&
$S[0,\infty)$&
$\mathbb{E}1, \mathbb{E}X^2, \mathbb{E}\ln X$&
$e^{\lambda_0-1} x^{\lambda_1} e^{\lambda_2 {x^2}}\mathbb{I}_{x\geq 0} $\\
\hline
Weibull&
$S[0,\infty)$&
$\mathbb{E}1, \mathbb{E}\ln X, \mathbb{E}X^a$
&$e^{\lambda_0} x^{\lambda_1} e^{\lambda_2 {x^a}}\mathbb{I}_{x\geq 0},~a>0~constant$\\
\hline
GED&
$S(-\infty,\infty)$&
$\mathbb{E}1, \mathbb{E}|X-\mu|^{\beta}$&
$e^{\lambda_0 -1} e^{\lambda_1|x-\mu|^{\beta}}$\\
\hline
Standard Cauchy&
$S(-\infty,\infty)$&$\mathbb{E}1, \mathbb{E}\ln (1+X^2)$
&$e^{\lambda_0 -1} (1+x^2)^{\lambda_1}$\\
\hline
Student's $t$&
$S(-\infty,\infty)$&
$\mathbb{E}1, \mathbb{E}\ln (1+{\nu}^{-1}X^2)$&
$e^{\lambda_0 -1} (1+{\nu}^{-1}x^2)^{\lambda_1}$\\
\hline
Pearson type-IV&
$S(-\infty,\infty)$&
$\mathbb{E}1, \mathbb{E}\ln ( 1+(\frac{X-\lambda}{\alpha})^2 ), \mathbb{E} \arctan(\frac{X-\lambda}{\alpha}) $&
$e^{\lambda_0 -1} [ 1+(\frac{x-\lambda}{\alpha})^2 ]^{\lambda_1} \exp[\lambda_2 \arctan(\frac{x-\lambda}{\alpha})]$\\
\hline
Generalized Student's $t$ \cite{park2009maximum}&
$S(-\infty,\infty)$&
$\mathbb{E}1, \mathbb{E}\ln (1+{\nu}^{-1}X^2)$&
$e^{\lambda_0 -1} (1+{\nu}^{-1}x^2)^{\lambda_1} \exp[\lambda_2 \arctan(\frac{x}{\alpha})] \exp\left(\sum_{i=3}^{M} \lambda_i x^{i-2} \right)$
\\
&
&$\mathbb{E} \arctan(\frac{X}{\alpha}), \mathbb{E} X^{i-2}, i=3,\cdots, M$
&
\\
\hline
Generalized log-normal \cite{park2009maximum}&
$S(0,\infty)$&
$\mathbb{E}1, \mathbb{E}\ln X, \mathbb{E}(\ln X)^2$
&$e^{\lambda_0-1}x^{\lambda_1} \exp\left[\lambda_2 (\ln x)^2\right] \exp\left(\sum_{i=3}^{M} \lambda_i x^{i-2} \right) \mathbb{I}_{x>0} $
\\
&
& $\mathbb{E} X^{i-2}, i=3,\cdots, M$
&
\\
\hline
L\'{e}vy \cite{tsallis1995statistical}&
$\frac{1}{1-q}\left(1-\int d(\frac{x}{\sigma})(\sigma p)^q\right)$&
$\mathbb{E}1$&
$\frac{1}{\sigma} \left(\frac{q}{\lambda_0 (q-1)} \right)^{\frac{1}{1-q}} \left(1-\lambda_1(q-1)x^2 \right)^{\frac{1}{1-q}}$\\
&
$ q \in \mathbb{R},\quad \sigma >0$& 
$\int d(\frac{x}{\sigma}) (\sigma p)^{q} x^2 = \sigma^2$
&\\
\hline
Stretched Exp. \cite{anteneodo1999maximum}&
$\int dx\bigg[ \Gamma(\frac{\eta+1}{\eta}, -\ln p)$&
$ \mathbb{E}1 $&
$\exp\left\{-\left[\lambda_0 + \sum_{i=1}^{M}\lambda_i f_{i}(x) + \Gamma(\frac{\eta+1}{\eta})\right]^{\eta} \right\}$\\
&
$-p\Gamma(\frac{\eta+1}{\eta}) \bigg],~ \eta >0$&
$\mathbb{E}({f}_{i}(x)) =  \langle {f}_{i}(X) \rangle, \quad i=1,\cdots, M$
&
&
\\
\hline
\end{tabular}
\end{table}

\begin{table}[]
\setlength\extrarowheight{12pt}
\setlength{\tabcolsep}{2pt}
\centering
\caption{Discrete optimal probabilities corresponding to various entropy functions and constraints}
\label{Tab2}
\begin{tabular}{llllllll}
\hline
Distribution& Entropy& Constraints& Optimal Distribution&\\
\hline
Discrete uniform&
$S\{a,a+1,\cdots,b\}$&
$\mathbb{E}1$&
$\frac{1}{b-a+1}$\\
\hline
Maxwell-Boltzmann&
$S\{1,2,\cdots,N\}$&
$\mathbb{E}1, \mathbb{E}X$&
$\frac{1}{\sum_{i=1}^N \exp( \lambda_1 x_i)}\exp( \lambda_1 x_i)$\\
\hline
Bernoulli&
$S\{0,1\}$&
$\mathbb{E}1, \mathbb{E}X$&
$P(X=1)=\mathbb{E}X, P(X=0)=1-\mathbb{E}X$
\\
\hline
Binomial \cite{harremoes2001binomial}&
$S\{0,1,\cdots,N\}$&
$\mathbb{E}X, B_N(\lambda)$&
$C_N^k (\frac{\lambda}{N})^k (\frac{\lambda}{N})^{(N-k)}$
\\
\hline
Poisson\cite{harremoes2001binomial}&
$S\{0,1,\cdots,\infty\}$&
$\mathbb{E}X, B_{\infty}(\lambda)$&
$\frac{{\lambda}^k e^{-\lambda}}{k!}$
\\
\hline
Geometric&
$S\{1,2,\cdots,\infty\}$&
$\mathbb{E}1,\mathbb{E}X$&
$\left(1-{\langle X\rangle}^{-1}\right)^{i-1}{\langle X\rangle}^{-1}$\\
\hline
\end{tabular}
\end{table}

\section{Maximum EntroPy Neural Network (MEP-Net)}
\label{The Architecture of MEP-Net}
\subsection{Basic Architecture}
To provide an efficient numerical implementation of the maximum entropy principle, we propose the MEP-Net to automatically infer the desired unknown probability distribution function from limited observable data. A key step of the MEP-Net is to adopt a deep neural network to approximate the distribution features of the data based on the universal approximation theorem of neural networks\cite{}.

\begin{figure}[]
\centering
\includegraphics[width=0.85\linewidth]{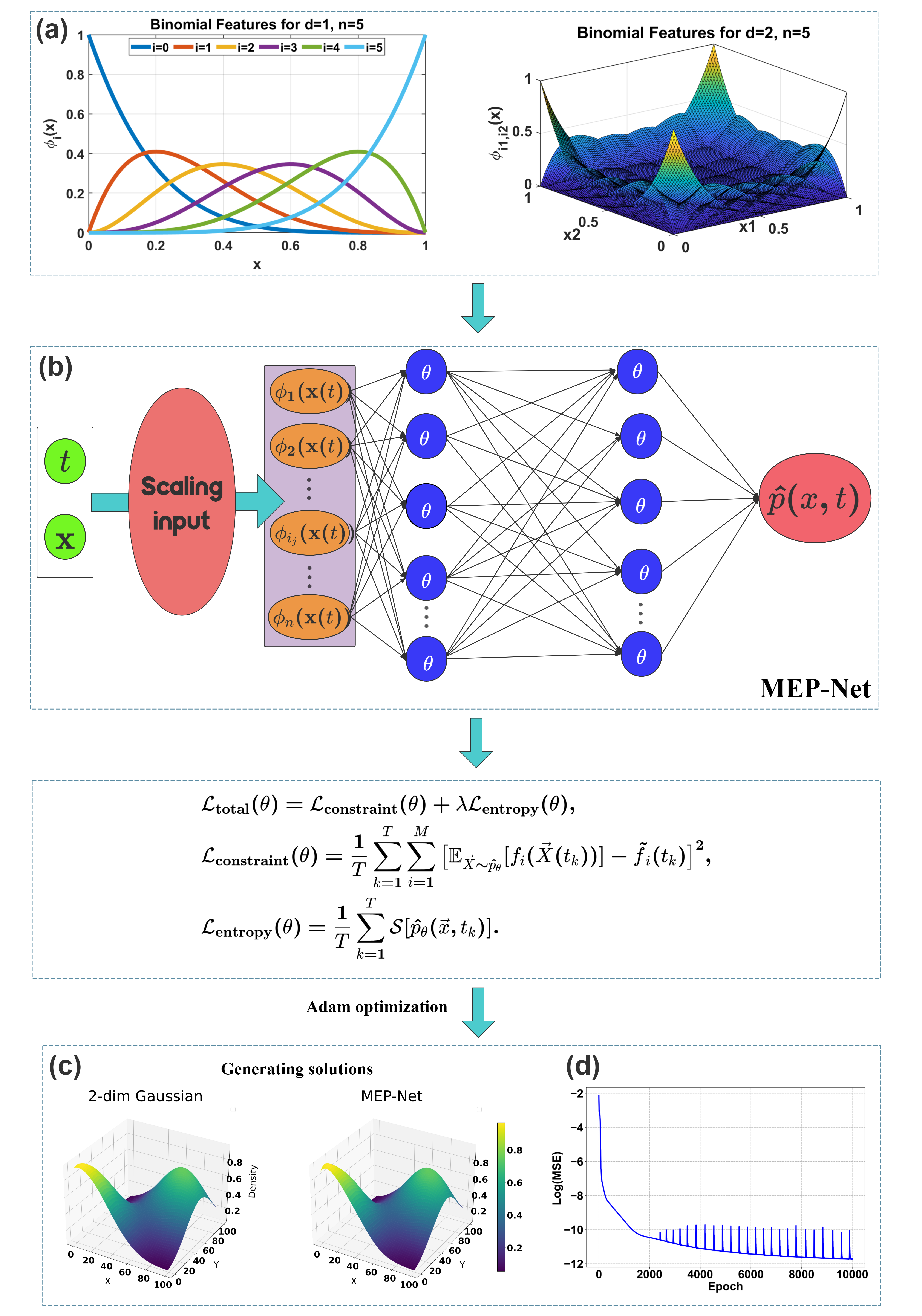}
\caption{\textbf{A schematic diagram for the MEP-Net.} (a) Generation of binomial features \(\phi_{i_1,i_2,\ldots,i_d}(\vec{x})\) based on data points. (b) Architecture of the MEP-Net. The inputs \(\vec{x}\) and \(t\) are scaled and then fed into the neural network.
The output is the generated probability distribution \(\hat{p}(\vec{x},t)\). (c) and (d) illustrate the generated probability distribution by the MEP-Net compared with the true one, and the evolution of the MSE during the training procedure.}
\label{fig.MaxEP-net}
\end{figure}

As illustrated in Fig. \ref{fig.MaxEP-net}, the loss function for the MEP-Net is given by
\begin{equation}
\mathcal{L}_{\text {total}}(\theta)=\mathcal{L}_{\text {constraint}}(\theta)+\lambda\mathcal{L}_{\text {entropy}}(\theta),
\end{equation}
where
\begin{equation*}
\begin{aligned}
&\mathcal{L}_{\text {constraint}}(\theta) = \frac1{T}\sum_{k=1}^T\sum_{i=1}^M \big[\mathbb{E}_{\vec{X}\sim \hat{p}_\theta}[f_i(\vec{X}(t_k))]-\tilde{f}_i(t_k)\big]^2, \\
&\mathcal{L}_{\text {entropy}}(\theta)=\frac{1}{T}\sum_{k=1}^T\mathcal{S}[\hat{p}_\theta(\vec{x},t_k)].
\end{aligned}
\end{equation*}
Here $\lambda >0$ is the weight, $\tilde{f}_i(t_k)$ is the data constraint at time $t_k$, and $\mathbb{E}_{\vec{X}\sim \hat{p}_\theta}[f_i(\vec{X}(t_k))] = \int_{\mathbb{R}^d} f_i(\vec{x}) \hat{p}_\theta(\vec{x},t_k) d\vec{x}$. $\mathcal{S}[\hat{p}_\theta(\vec{x},t_k)]$ denotes the entropy function as a functional of the output of the neural network $\hat p_{\theta}(\vec{x},t_k)$. 
Despite of its significance, the concrete form of the entropy function that fits the desired probability distribution is generally unavailable. 

To solve this difficulty, we propose a practical form based on the concept of fixed-point iteration,
\begin{equation}
\label{fixed-point-entropy}
\mathcal{S}[\hat{p}_\theta(\vec{x},t_k)]=\int_{\mathbb{R}^d}d\vec{x}\hat{p}_\theta(\vec{x},t_k)\ln\left[\frac{\hat{p}_\theta(\vec{x},t_k)}{\hat{p}_\theta(\vec{x},t_k)_{old}}\right],
\end{equation}
where $\hat{p}_\theta(\vec{x},t_k)_{old}$ represents the predicted probability distribution by the MEP-Net in the previous iteration. Based on the Kullback-Leibler divergence, it is straightforward to see that $\mathcal{S}[\hat{p}_\theta(\vec{x},t_k)]$ is always non-negative and reaches its minimum if and only if $\hat{p}_\theta(\vec{x},t_k)=\hat{p}_\theta(\vec{x},t_k)_{old}$, meaning the output of the MEP-Net converges. Therefore, we have two opposite contributions: the constraint loss forces the MEP-Net to approach the one best fitting the data constraints as soon as possible, while the entropy loss aims to keep the changes in the MEP-Net slow and smooth. 



\subsection{Binomial Functions as Observables}
The quality of data and the form of constraints determine the performance of the MEP-NEt in a significant way. Here we make an exploration on the reconstruction ability of various nonlinear features (as constraints), including polynomial, exponential, logarithmic, sine, cosine, square root and reciprocal functions, as well as their combinations. 
Unfortunately, none of them could achieve a satisfactory reconstruction of the desired probability distributions under test.

Inspired by the observations from the Refs.\cite{barzel2011binomial,jia2020kinetic}, we adopt the binomial functions instead, which is defined as
\begin{equation}
\phi_{k_1,k_2,\ldots,k_d}(\vec{x})=\prod_{j=1}^d\binom n{k_j}x_j^{k_j}(1-x_j)^{n-k_j},
\end{equation}
where $k_j \in \{0, 1, \dots, n\}$ represents the power of the $j$-th dimension, and the combinatorial number $\binom{n}{k_j}=\frac{n!}{k_j!(n-k_j)!}$ gives the binomial coefficient. The binomial functions, due to their combinatorial properties, can well capture the complex interrelationships among variables. As show in Fig. \ref{fig.gauss}, the adoption of the binomial functions as observables (constraints) yields unexpectedly positive results. Even those hard tasks, like the multimodal distributions and the Beta distribution, could be easily reconstructed in a very precision based on only a few terms of the binomial functions.

To sum up, our design of the MEP-Net incorporates several key features that significantly enhance its performance and applicability. Firstly, the model employs binomial features to boost its expressive power, enabling it to capture complex data distributions in a better way. Secondly, an entropy loss is introduced as a regularization term, effectively maintaining the model's output and improving its generalization capability. Furthermore, the incorporation of the maximum entropy principle ensures that the generated probability distribution is the most unbiased one in physics, thereby guaranteeing the physical plausibility of the results. Lastly, the MEP-Net cleverly applies the concept of fixed-point iteration, by utilizing the predictions from the previous iteration to optimize the entropy loss. This approach not only improves the computational efficiency but also enhances the stability of the model. Overall, the combination of the above features enables the MEP-Net to excel in generating probability distributions with limited knowledge for scientific problems. 

\subsection{Computational Setup}
In our applications, the MEP-Net includes two hidden layers with 50 nodes each, followed by an output layer with a single output. Only the first hidden layer employs a Tanh activation function, while the output layer adopts a Softplus activation function, facilitating non-linear transformations of the input data. We also explore other configurations of neural networks, but none matches the effectiveness of this particular architecture.

All experiments are conducted using a hardware configuration consisting of 1 NVIDIA RTX 4090D GPU with 24GB VRAM, 16 vCPU Intel(R) Xeon(R) Platinum 8474C processor, and 80GB RAM. The software environment includes Python version 3.8, PyTorch version 1.11.0, and CUDA version 11.3. This setup provides the necessary computational power for training our MEP-Net models and running the numerical experiments described in the following sections.

\section{Numerical Experiments}
\label{Numerical Experiments}
In this section, we will present a number of examples to comprehensively demonstrate the performance and application potential of the MEP-Net. First, we evaluate the expression ability of the MEP-Net in generating various statistical distributions, including the Gaussian mixtures, Beta distribution, high-dimensional distributions and time-dependent Gaussian functions. Then, we demonstrate that the MEP-Net can accurately generate the solutions of the Schlögl model, which is a classic chemical reaction model exhibiting rich dynamical behaviors, such as the bi-stability and stochastic transitions between stable states, revealing its potential in handling complex chemical reaction systems. Next, we utilize the MEP-Net to solve variational problems in non-equilibrium systems, the diffusion equation formulated through the Onsager's variational principle to be exact\cite{doi2011onsager}. 
In the last example, we apply the MEP-NEt to the dynamical procedure of phase separation described by the Allen-Cahn equation, which has wide applications in fields such as material science and biology science.  

\subsection{Statistical Distributions}
\subsubsection{One-dimensional Case}
\label{One-dimensional Case}
Consider the general form of a multimodal Gaussian distribution $p(x)$ for a one-dimensional random variable $X$,
\begin{equation}
p(x)=\sum_{i=1}^Nw_i\mathcal{N}(x;\mu_i,\sigma_i^2), \quad \mathcal{N}(x;\mu_i,\sigma_i^2)=\frac1{\sqrt{2\pi\sigma_i^2}}\exp\left(-\frac{(x-\mu_i)^2}{2\sigma_i^2}\right),
\end{equation}
where the coefficient $w_i \geq 0$ denotes the weight of the $i$-th Gaussian distribution satisfying $\sum_{i=1}^N w_i=1$, $N$ is the number of Gaussian distributions. Each Gaussian is characterized by its mean $\mu_i$ and variance $\sigma_i^2$. 

\begin{figure}[]
\centering
\includegraphics[width=0.85\linewidth]{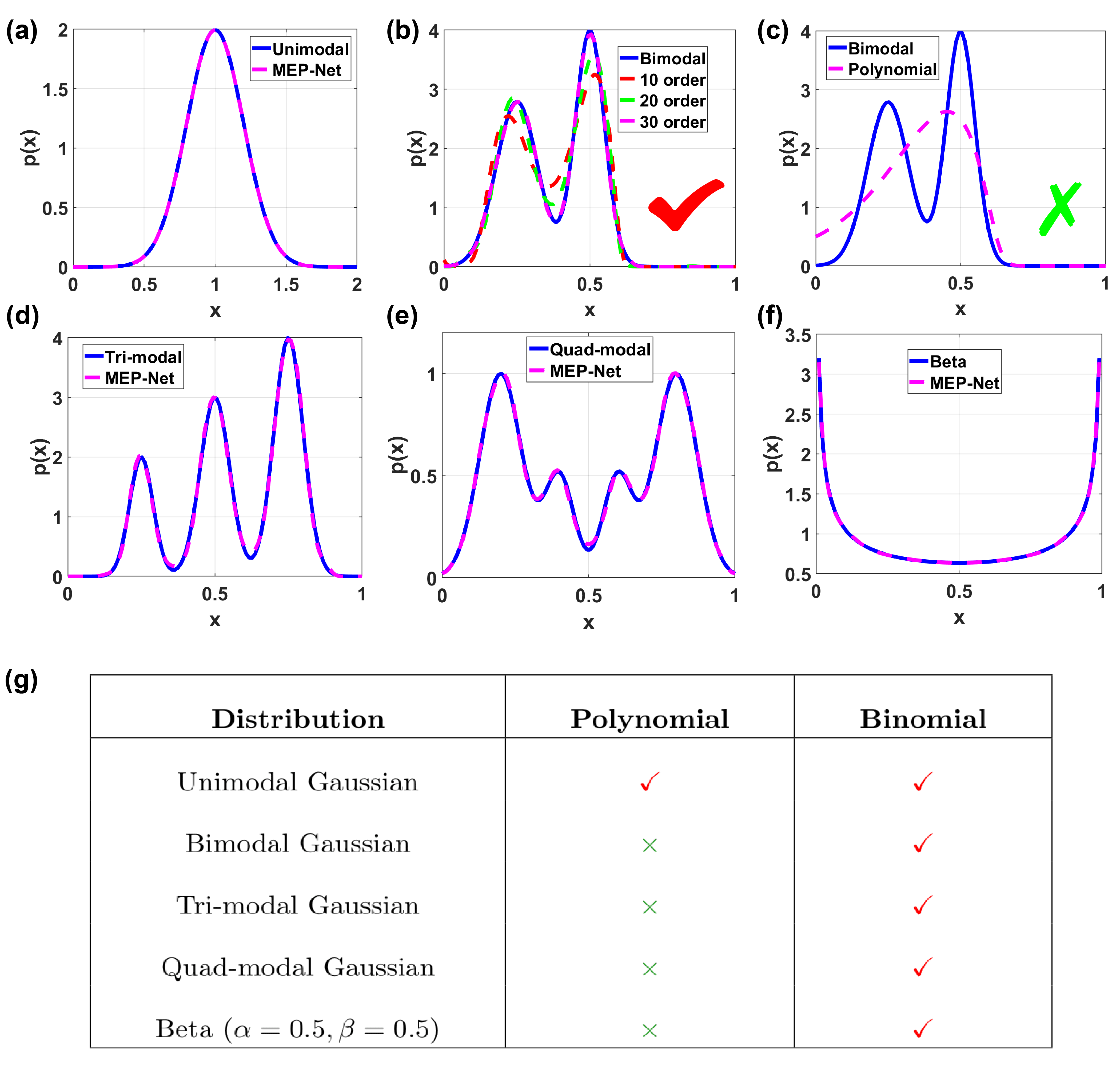}
\caption{\textbf{Generation of one-dimensional Gaussian mixture distributions by MEP-Net.} (a) Unimodal Gaussian distribution with $\mu_1=1$ and $\sigma_1=\displaystyle\frac1{5}$. (b) Bimodal Gaussian distribution with $\mu_1=\displaystyle\frac14,\sigma_1=\displaystyle\frac1{14}$ for the first mode and $\mu_2=\displaystyle\frac12,\sigma_2=\displaystyle\frac1{20}$ for the second mode. (c) Reconstruction of a Bimodal Gaussian distribution by polynomial moments. (d) Tri-modal Gaussian distribution with $\mu_1=\displaystyle\frac14,\sigma_1=\displaystyle\frac1{25}$ for the first mode, $\mu_2=\displaystyle\frac12,\sigma_2=\displaystyle\frac1{20}$ and $\mu_3=\displaystyle\frac34,\sigma_3=\displaystyle\frac1{20}$ for the second and third modes. (e) Quad-modal Gaussian distribution with means at $\displaystyle\frac15,\displaystyle\frac25,\displaystyle\frac35$, and $\displaystyle\frac 45$, standard deviations of $\displaystyle\frac1{14},\displaystyle\frac1{20},\displaystyle\frac1{20}$, and $\displaystyle\frac1{14}$. (f) Beta distribution, $\alpha=\beta=\displaystyle\frac12$. (g) Comparison on the reconstruction abilities of different types of moments. Checkmarks (\(\checkmark\)) indicate those successful generations, while crosses (\(\times\)) indicate the unsuccessful ones.}
\label{fig.gauss}
\end{figure}

As illustrated in Fig. \ref{fig.gauss}, the MEP-Net is capable of reconstructing various multimodal Gaussian distributions, ranging from uni-modal to quad-modal distributions. The MEP-Net accurately captures and reproduces the exact positions of all peaks, their width and relative heights. Particularly, in Fig. \ref{fig.gauss}(b) we make a detailed comparison on the reconstruction abilities of binomial functions with 10, 20, and 30 terms respectively. It is evident that the inclusion of more terms leads to an improved representation of the bimodal characteristics. In Fig. \ref{fig.gauss}(f), we successfully reconstruct the Beta distribution as well, whose form is given by $p(x;\alpha,\beta)=x^{\alpha-1}(1-x)^{\beta-1}$. 

\subsubsection{High-dimensional Case}
\label{High-dimensional Case}
We proceed to the high-dimensional distributions. Let $\vec{X}\in \mathbb{R}^d$ be a $d$-dimensional random vector. 
The pdf of the Gaussian mixture is given by
\begin{equation}
\begin{aligned}
&p(\vec{x}; \boldsymbol{\Theta}) = \sum_{i=1}^{N} w_i \mathcal{N}(\vec{x};\vec{\mu}_i,\boldsymbol{\Sigma}_i), \\
&\mathcal{N}(\vec{x}; \vec{\mu}_i,\boldsymbol{\Sigma}_i) = \frac{1}{(2\pi)^{d/2}|\boldsymbol{\Sigma}_i|^{1/2}}\exp\left(-\frac{1}{2}(\vec{x}-\vec{\mu}_i)^T\boldsymbol{\Sigma}_i^{-1}(\vec{x}-\vec{\mu}_i)\right),
\end{aligned}
\end{equation}
where $w_i \geq 0$ denotes the weight of the $i$-th Gaussian distribution satisfying $\sum_{i=1}^{N} w_i=1$, and $\boldsymbol{\Theta} = \{w_1,\dots,w_N,\vec{\mu}_1,\dots,\vec{\mu}_N,\boldsymbol{\Sigma}_1,\dots,\boldsymbol{\Sigma}_N\}$ denotes the set of model parameters. 
$\mathcal{N}(\vec{x};\vec{\mu}_i,\boldsymbol{\Sigma}_i)$ represents the $d$-dimensional multivariate normal distribution, in which $\vec{\mu}_i \in \mathbb{R}^d$ is the mean vector, $\boldsymbol{\Sigma}_i \in \mathbb{R}^{d\times d}$ is the positive definite covariance matrix, $|\boldsymbol{\Sigma}_i|$ is the determinant, and  $\boldsymbol{\Sigma}_i^{-1}$ is the inverse of the $i$-th covariance matrix.

\begin{figure}
\centering
\includegraphics[width=1.0\linewidth]{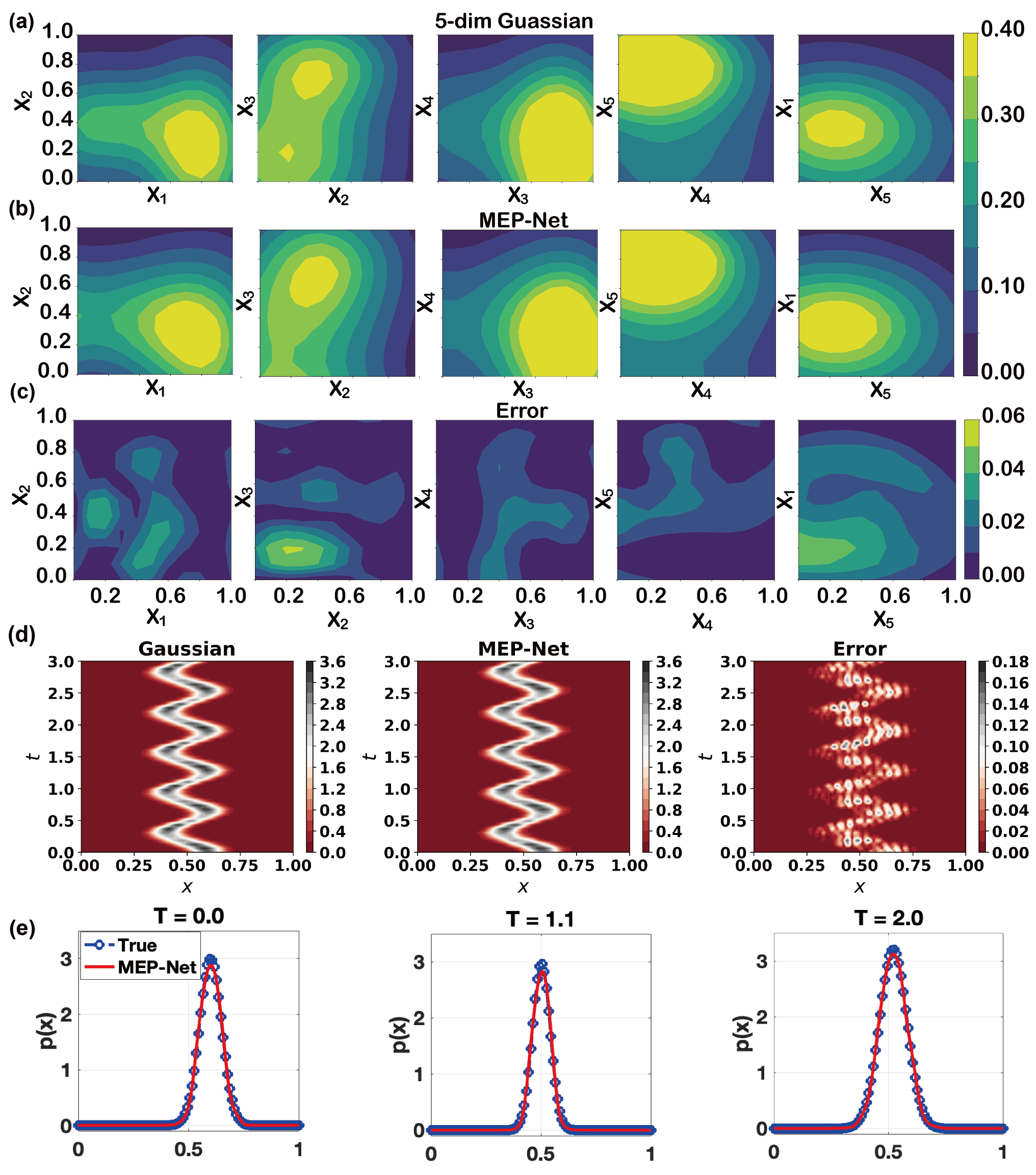}
\caption{\textbf{5-dimensional Gaussian mixture distributions and time-dependent Gaussian function.} Comparisons on (a) the true and (b) generated 5-dimensional Gaussian mixture distributions by the MEP-Net, with the contour plots for $(x_1, x_2), (x_2, x_3), (x_3, x_4), (x_4, x_5)$, and $(x_5, x_1)$ listed from left to right. Their absolute errors are given in (c).  (d) The true values of time-dependent Gaussian function (left panel), the predicted values by the MEP-Net (middle panel), and their difference (right panel). (e) Snapshots of the time-dependent Gaussian function at $t=0$, $t=1.1$ and $t=2.0$, respectively.}
\label{fig.High-dimensional}
\end{figure}

We test higher-dimensional examples by considering a 5-dimensional Gaussian mixture. It consists of 3 components, with $w_1 = 0.4$, $w_2 = 0.3$, and $w_3 = 0.3$ as the mixing weights, 
$\vec{\mu}_1=(0.05,0.15,0.05,0.15,0.05)^T$, $\vec{\mu}_2=(0.2,0.4,0.2,0.8,0.2)^T$, $\vec{\mu}_3=(0.8,0.2,0.4,0.2,0.4)^T$
being the mean vectors, and $\boldsymbol{\Sigma}_1, \boldsymbol{\Sigma}_2, \boldsymbol{\Sigma}_3$ being the diagonal covariance matrices, i.e.
$\boldsymbol{\Sigma}_{1}=diag(0.1, 0.4, 0.2, 0.3, 0.15)$,
$\boldsymbol{\Sigma}_{2}=diag(0.15, 0.05, 0.15, 0.05, 0.15)$,
$\boldsymbol{\Sigma}_{3}=diag(0.05, 0.15, 0.1, 0.2, 0.25)$.
In this case, the absolute errors between the MEP-NET predictions and the true 5-dimensional distribution could be maintained less than $6\%$ (see Figs. \ref{fig.High-dimensional}(a-c)).

\subsubsection{Time-Dependent Gaussian Function}
We extend our application to a time-dependent Gaussian function
\begin{equation*}
p(x,t)=A(t)\exp\left(-\frac{(x-\mu(t))^2}{2\sigma(t)^2}\right),
\end{equation*}
whose amplitude, mean, and standard deviation evolve over time according to  $A(t)=3+0.5\sin(20t)$, $\mu(t) =0.5+0.1\cos(10 t)$ and $\sigma\left(t\right) =0.05+0.01\sin(10 t)$ respectively. Notice that this Gaussian function is not normalized to be a probability density function. The reconstruction results for this time-varying Gaussian function are illustrated in Figs. \ref{fig.High-dimensional}(d,e), highlighting the MEP-Net's promising capability to capture and reproduce the temporal variations inherent in such functions.

\subsection{Chemical Master Equations}
\begin{figure}[]
\centering
\includegraphics[width=1.0\linewidth]{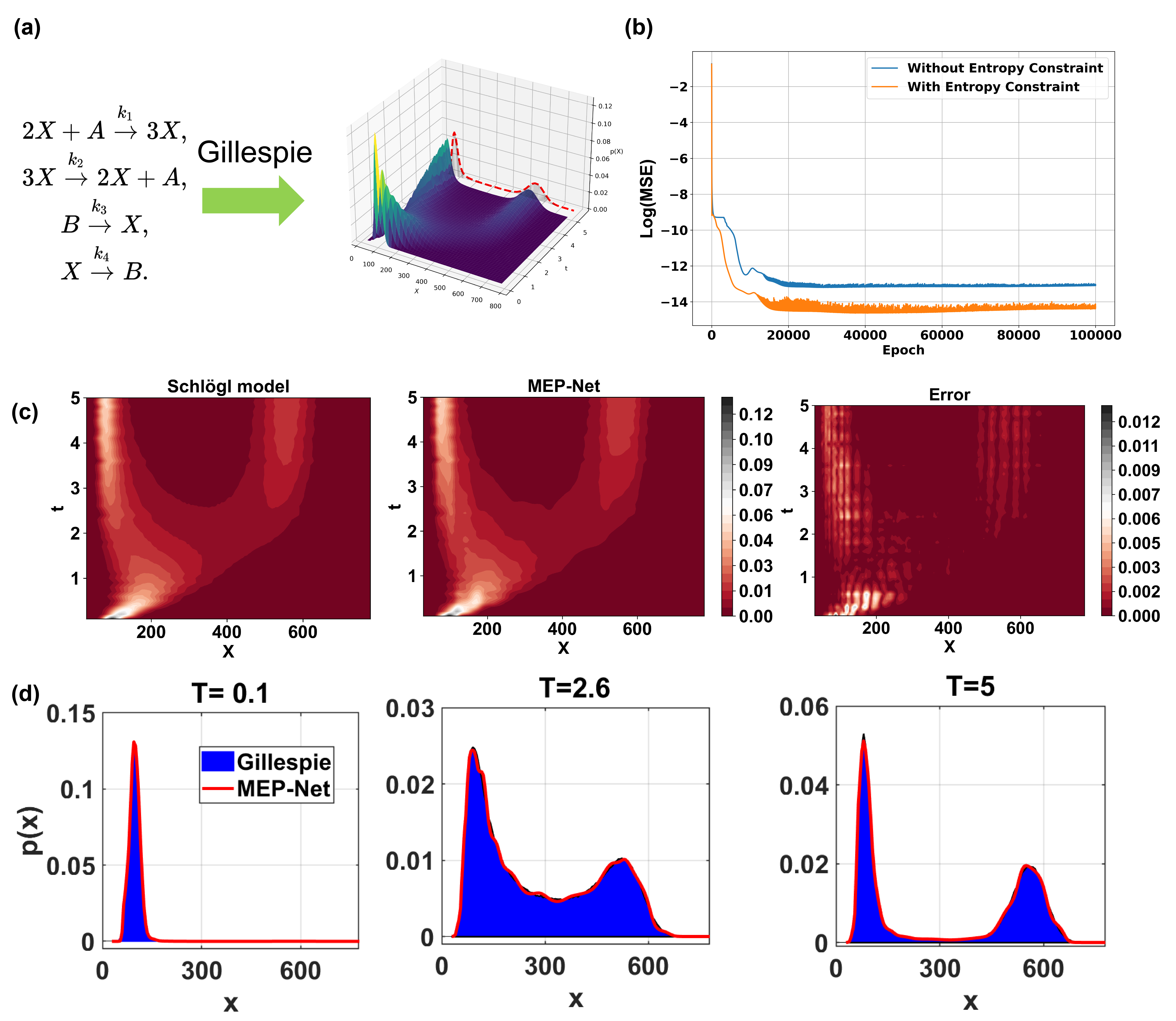}
\caption{\textbf{Dynamics of the Schlögl model and generation by the MEP-Net.} (a) The mechanism of the Schlögl model and data generated via the Gillespie algorithm. (b) Logarithmic Mean Squared Errors (MSE) of the MEP-Net with and without the  entropy loss as a function of the training epoch. (c) Heatmaps for the dynamics predicted by the Schlögl model (left) and by the MEP-Net (center), with their absolute errors shown on the right. (d) Snapshots of the pdfs of the Schlögl model and the MEP-Net generated at $T=0.1$, $T=2.6$, and $T=5.0$, respectively.}
\label{fig.vartr}
\end{figure}

The next example we studied is the Schlögl model\cite{}, which consists of a set of reversible auto-catalytic reactions involving three substances $X$, $A$, and $B$ (see Fig. \ref{fig.vartr}(a)). 
Here the molecule numbers of $A$ and $B$ are kept unchanged. $p(n,t)$ represents the probability of observing $n$ molecules of $X$ at time $t$. Based on laws of mass-action, the chemical master equation reads, 
\begin{equation}
\begin{aligned}
\frac{dp(n,t)}{dt}&=\left[k_{1}'(n-1)(n-2)+k_{3}'\right]p(n-1,t)
-\left[k_1'n(n-1)+k_3'\right]p(n,t) \\
&+\left[k_{2}(n+1)n(n-1)+k_{4}(n+1)\right]p(n+1,t) 
-\left[k_2n(n-1)(n-2)+k_4n\right]p(n,t). 
\end{aligned}
\end{equation}
where $k_1'=k_1[A], k_3'=k_3[B]$ represent two pseudo reaction rate constants. 

A notable feature of the Schlögl model is that under certain conditions, the molecular number of $X$ can exhibit bi-stability, characterized by two distinct stable states\cite{smadbeck2013closure,rathinam2005consistency}. To examine this phenomenon, we choose $k_1 = 1.5 \times 10^{-7}, k_2=1.0 \times 10^{-4} / 6, k_3=1.0 \times 10^{-3}, k_4 = 3.5$. The initial condition is set as $[X](0) = 250, [A](0) = 1.0 \times 10^{5}, [B](0) = 2.0 \times 10^{5}$. The Gillespie's algorithm is implemented to generate one million independent trajectories to get a probability distribution \cite{gillespie1977exact}. 


The bi-stability of the Schlögl model is clearly demonstrated through the bimodal distributions exhibited in Figs. \ref{fig.vartr}(c-d). And the MEP-Net can reach a remarkable accuracy in generating the time-varying solutions to the Schlögl model, with the absolute errors below $1.2\%$. Furthermore, the MSE of the MEP-Net is about 1-2 orders of magnitude lower than the one without the entropy loss (see Fig. \ref{fig.vartr}(b)), suggesting the usefulness of including the entropy loss in the MEP-Net.

\subsection{Particle Diffusion in a Confined Domain}
The architecture of the MEP-Net is particularly suitable for solving various scientific problems formulated in a variational form. Let us consider Brownian particles confined by two walls located at $x=-a$ and $x=a$. The governing equation for particle density $\eta=\eta(x,t)$ reads \cite{doi2015onsager} 
\begin{equation}
\label{diffusion eq}
\begin{aligned}
&\frac{\partial \eta(x,t)}{\partial t}=D\frac{\partial^2\eta(x,t)}{\partial x^2},\\
&\eta(x,0)=\eta_0+\eta_1\mathrm{sign}(x),\quad 
 \left.\frac{\partial \eta}{\partial x}\right|_{x=-a}= \left.\frac{\partial \eta}{\partial x}\right|_{x=a}=0,
\end{aligned}
\end{equation}
where the diffusion constant is related to the viscosity coefficient through the Einstein's relation $D=k_\text{B}T/\zeta$, $k_B$ is the Boltzmann constant and $T$ is the temperature. The above equation can easily solved by using Fourier transforms, which leads to a series solution,
\begin{equation}
\eta(x,t)=\eta_0+\eta_1\sum_{p=0}^\infty c_p\sin(\lambda_p x)e^{-\mu_p t},
\end{equation}
where the coefficients $\lambda_p=\displaystyle\frac{(2p+1)\pi}{2a}$, $\mu_p=\displaystyle\frac{(2p+1)^2\pi^2}{4a^2}{D}$ and $c_p=\displaystyle\frac{4}{(2p+1)\pi}$. 

Doi has pointed out that the diffusion equation could be reformulated into a variational form by using the so-called Onsager's variational approach \cite{doi2015onsager}. In this approach, the key issue is to construct an appropriate variational function -- the Rayleighian $\mathcal{R}$, which is defined as the summation of the dissipation function $\mathcal{D}$ and the time derivative of the free energy $\dot{\mathcal{F}}$,
\begin{equation}
\label{diffusion-entropy}
\mathcal{R} = \mathcal{D} + \dot{\mathcal{F}}. 
\end{equation}
In this case, the dissipation function $\mathcal{D}$ is related to the flux velocity of particles $v(x,t)$, which can be expressed as,
\begin{equation*}
\mathcal{D} = \frac{1}{2} \int dx\zeta \eta(x,t)v(x,t)^2.
\end{equation*}
On the other hand, the free energy $\mathcal{F}$ of this system can be written as a functional of $\eta(x,t)$:
\begin{equation*}
\mathcal{F}[\eta(x,t)] = \int dxk_BT [\eta(x,t) \ln \eta(x,t)].
\end{equation*}
By minimizing the Rayleighian with respect to the flux velocity $v$
\begin{equation}
\min_{v(x,t)} \int dt \left[ \mathcal{D} + \dot{\mathcal{F}}\right]=\min_{v(x,t)} \int dt \left[ \frac{1}{2} \int dx\zeta \eta(x,t)v(x,t)^2+\int dxk_BT\frac{\partial\eta(x,t)}{\partial t} \ln 
\eta(x,t)\right],
\end{equation}
and utilizing the continuity equation $\partial \eta(x,t)/\partial t+\partial(\eta(x,t)v(x,t))/\partial x=0$, we can obtain the evolution equation for the particle density $\eta(x,t)$ in Eq. \eqref{diffusion eq}. 

\begin{figure}[]
\centering
\includegraphics[width=1.0\linewidth]{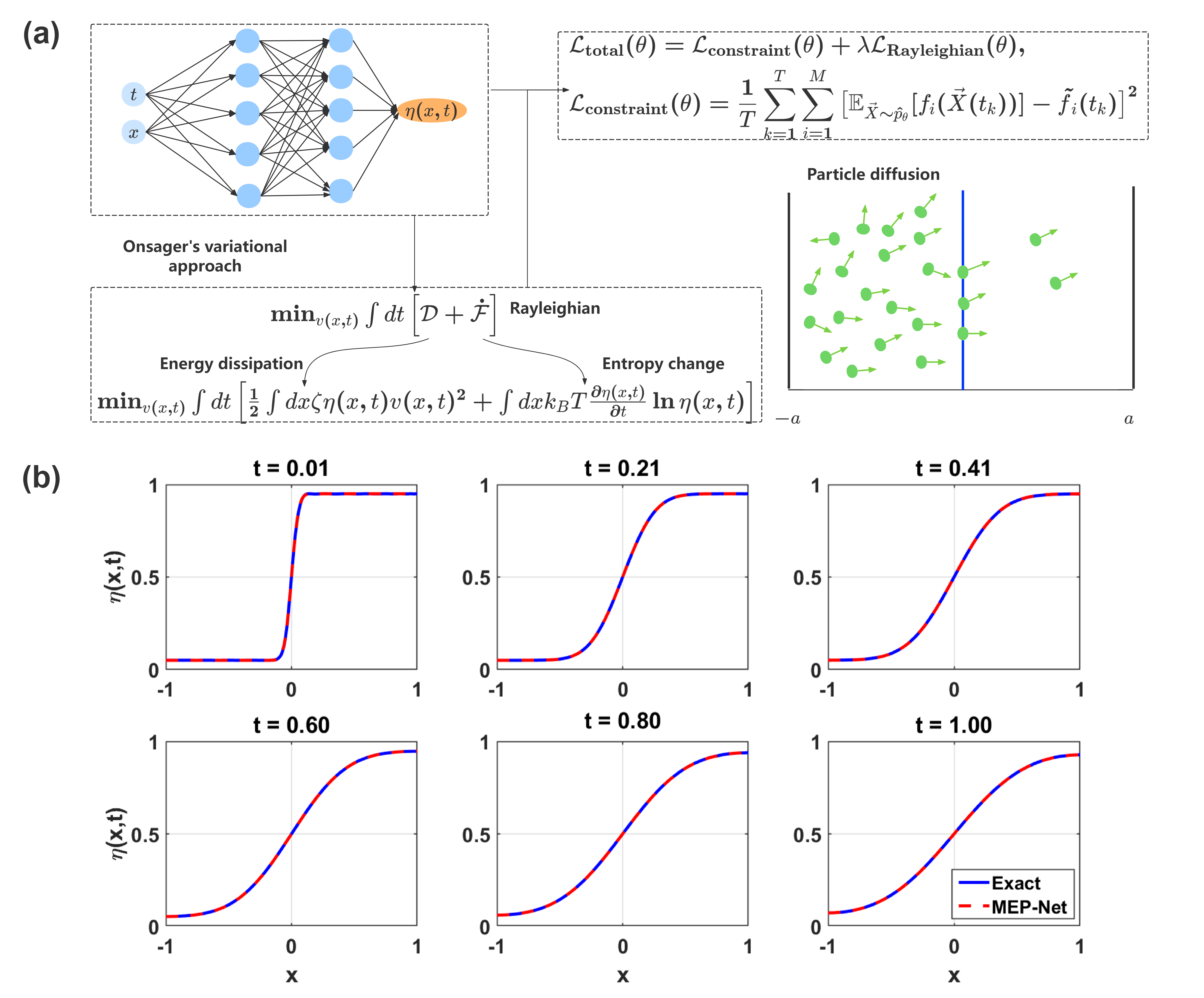}
\caption{\textbf{Particle diffusion in a confined domain}. (a) Architecture of the MEP-Net with the entropy loss being replaced by the Rayleighian. (b) Comparison between the approximate solution obtained by the MEP-Net (represented by the dashed line) and the exact solution (represented by the blue solid line) at time instances t = 0.01, 0.21, 0.41, 0.60, 0.80, and 1.00. The spatial domain is defined as $[-1, 1]$ with initial conditions $\eta_0 = 0.5$ and $\eta_1 = 0.45$, and the diffusion coefficient $D = 1.0$.}
\label{fig.diffusion}
\end{figure}

Due to the availability of the variational function, we will use the Rayleighian in Eq. \eqref{diffusion-entropy} in replace of the entropy loss in Eq. \eqref{fixed-point-entropy}. In this way, our MEP-Net turns to be a deep-learning-based solvers for the Onsager's variational approach, whose efficiency and accuracy is clearly demonstrated through the numerical experiments in Fig. \ref{fig.diffusion}(b).

\subsection{Allen-Cahn Model for Binodal Decomposition}
\label{Allen-Cahn Model for Binodal Decomposition}
The Allen-Cahn model provides a simple mechanism to understand the process of binodal decomposition, a typical phase separation phenomenon that occurs in plenty of material systems \cite{cahn1961spinodal}. The Allen-Cahn model is a special case of gradient flows \cite{fife1988dynamics}, which states the time evolution of the phase-field variable $\phi=\phi(\vec{x},t)$ satisfying the following gradient structure
\begin{eqnarray}
\label{gradient-flow}
\frac{\partial\phi}{\partial t}=-G\frac{\delta F[\phi]}{\delta\phi},
\end{eqnarray}
where $G$ is a positive definite operator, $\delta F[\phi]/\delta\phi$ is the variation of the free energy functional $F[\phi]$ with respect to $\phi$, the opposite of which points out the steepest descent direction of the free energy. A direct consequence of the gradient flow is the minimum free energy principle\cite{}, since we have
\begin{eqnarray*}
\frac{dF[\phi]}{dt}
=\int_{\Omega} d\vec{x}\frac{\delta F[\phi]}{\delta\phi}\frac{\partial\phi}{\partial t}
=-\int_{\Omega} d\vec{x}\frac{\delta F[\phi]}{\delta\phi}G\frac{\delta F[\phi]}{\delta\phi}\leq0, 
\end{eqnarray*}
where ${\Omega}$ is the region of interest. 

To derive the Allen-Cahn model, we specify the operator $G$ as an identity matrix ($G=I$), and the free energy functional
\begin{equation}
\label{ac-free-energy}
\mathcal{F}[\phi] = \int_{\Omega}\left[\frac{\varepsilon^2}{2}|\nabla \phi|^2 + f(\phi)\right] d\vec{x},
\end{equation}
where $\varepsilon\ll 1$ is a small parameter ($\varepsilon=0.025$ in the current case), and $f(\phi)$ is a double-well potential, typically given by $f(\phi) = \frac{1}{4}(\phi^2 - 1)^2$. Inserting them into Eq. \eqref{gradient-flow}, we arrive at the classical Allen-Cahn model, 
\begin{equation}
\frac{\partial\phi}{\partial t} = \varepsilon^2\Delta \phi - \phi^3 + \phi.
\end{equation}
The first term $\varepsilon^2\Delta\phi$ on the right-hand side represents the interfacial effect, which tends to smooth the interface. Meanwhile, the second and third terms drive the system towards stable states. 

 
Accompanied with the governing equation, we set the initial condition as
$$\phi(x,y,0)=\tanh\left(\frac{0.35-r}{2\times25}\right),$$
where $r=\sqrt{x^2+y^2}$ is the radial distance from the center of the domain. Meanwhile, we choose the Neumann boundary conditions, which assume that there is no flux across the boundaries of the domain. 
$$\frac{\partial \phi}{\partial \vec{n}} = 0, \quad\text{on } \partial \Omega.$$ 
Here $\partial \Omega$ represents the boundary of the domain $\Omega$, and $\vec{n}$ is the outward-pointing unit normal vector to the boundary. 

\begin{figure}[]
\centering
\includegraphics[width=\linewidth]{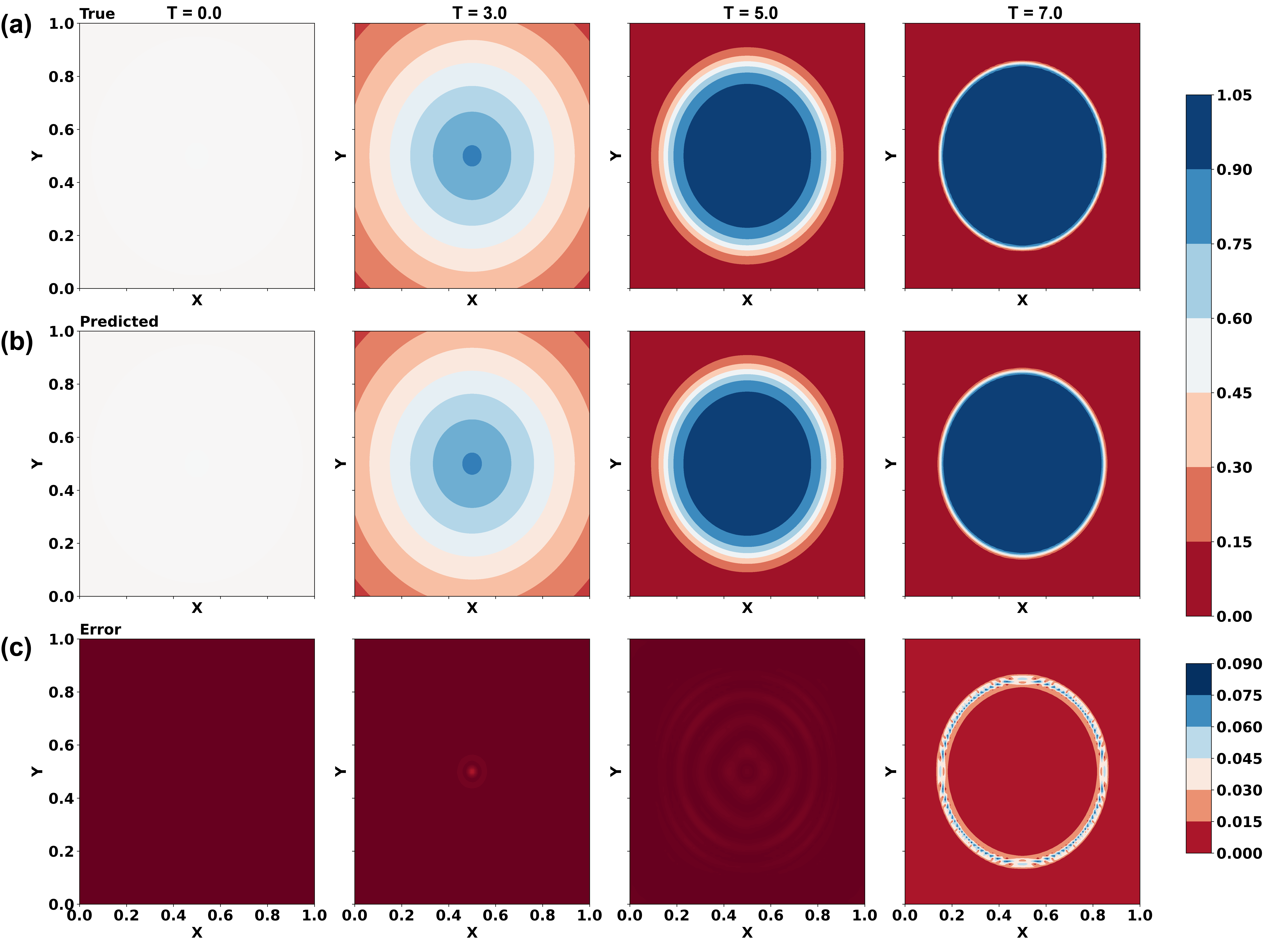}
\caption{\textbf{Allen-Cahn model for Binodal decomposition.} (a) The ground truth of the Allen-Cahn equation, showing the evolution of phase boundaries over time. (b) Solutions generated via the MEP-Net. (c) The absolute errors between the true values and the MEP-Net-generated solutions. The columns represent four different time steps $T=0.0$, $3.0$, $5.0$, and $7.0$, respectively.} 
\label{fig.Allen-CahnModel}
\end{figure}

To apply the MEP-Net to solve the Allen-Cahn model, we replace the entropy loss by the free energy in Eq. \eqref{ac-free-energy}. From Fig. \ref{fig.Allen-CahnModel}, we observe that MEP-Net can effectively generate solutions to the Allen-Cahn equation in a high precision. In particular, the MEP-Net shows a significant decrease in the mean squared error compared to the one without the entropy loss (see Fig. \ref{fig.AC_loss}).  

\section{Conclusion and Discussion}
\label{Conclusion and Discussion}
In this work, we have proposed the MEP-Net, a novel approach that combines the Maximum Entropy Principle with neural networks to generate probability distributions under limited information in the form of moment constraints. By introducing a practical entropy loss function in an iterative form and utilizing the binomial functions as data constraints, we have successfully solved two key problems during the numerical applications of the MEP.  And through fruitful numerical experiments, the outstanding performance of the MEP-Net for either generating multi-modal statistical distributions, or time-dependent probability solutions to chemical master equations, diffusion equations with a confined domain, and the Allen-Cahnn model for binodal decomposition, has been clearly demonstrated. In contrast, traditional methods for probability distribution reconstruction, such as statistical approaches and recent neural network-based generative models like GANs, VAEs, either struggle with limited data or require vast amounts of data to learn complex distributions. Our MEP-Net bridges this gap by leveraging the strengths of both MEP and neural networks. 

Future work on the MEP-Net holds promising directions for inverse problems and integration with other machine learning methods. In the domain of inverse problems, the MEP-Net's ability to generate physically meaningful probability distributions from limited information could be applied to image reconstruction, signal processing, and computational physics. This approach may yield improved solutions to long-standing problems in these fields, particularly when data is scarce or expensive to obtain.

Regarding to integration with other machine learning paradigms, combining the MEP-Net with those well-established approaches such as reinforcement learning or Bayesian inference could lead to the development of more robust and versatile hybrid models. These integrations may significantly enhance the MEP-Net's applicability to complex scenarios that benefit from multiple learning approaches, potentially expanding its utility in areas like decision-making under uncertainty or adaptive system modeling.

Furthermore, K.A. Dill \textit{et al.} developed a method called the MaxCal, which combines the Maximum Entropy Principle with thermodynamic model calibration, to reconstruct the energy landscapes and partition functions of dynamical systems from experimental data\cite{ge2012markov,Presse2013Principles,hazoglou2015communication}. How to incorporate the MaxCal with suitable neural network structures is of great interest.

\section*
{Declaration of Competing Interest}
Authors declare that they have no conflict of interest.

\section*
{Author Contributions}
\textbf{Wuyue Yang: } Conceptualization; Formal analysis;
Funding acquisition; Methodology; Validation; Visualization; Writing-original draft; Writing-review \& editing. \textbf{Liangrong Peng:} Conceptualization; Formal analysis; Funding acquisition; Methodology; Validation; Visualization; Writing-original draft;
Writing-review \& editing. \textbf{Guojie Li: } Conceptualization; Formal analysis;Validation; Visualization; Writing-review \& editing. \textbf{Liu Hong:} Conceptualization; Formal analysis; Funding acquisition; Methodology; Writing-original draft; Writing-review \& editing. 

\section*{Acknowledgements}
This work was supported by the National Key R\&D Program of China
(Grant No. 2023YFC2308702), the National Natural Science Foundation of China
(12205135,12301617), Guangdong Basic and Applied Basic Research Foundation
(2023A1515010157).

\section*{Data Availability Statement}
The data that support the findings of this study are available from the corresponding author upon request.

\appendix

\section{Rigorous derivations of various continuous distribution functions by MEP}
\label{Rigorous derivations of various continuous distribution}
\begin{prop}
\label{prop1}
For a $1$-d random variable $X$ with two observables, the mean $\langle X \rangle\equiv \mu$ and the variance $\langle (X  - \langle X \rangle)^2\rangle=\langle X^2 \rangle - \langle X \rangle^2 \equiv \sigma^2$. 
Then if and only if the probability density function is normally distributed with its mean as $\mu$ and its variance as $\sigma^2$, the Boltzmann-Gibbs (BG) entropy function  $S[p]=-\int_{-\infty}^{+\infty} dx \left(p(x) \ln p(x)\right)$ reaches its maximum value.
\end{prop}

\begin{proof}
We choose the classical BG entropy as the objective function $S[p]=-\int_{-\infty}^{+\infty} dx \left(p(x) \ln p(x)\right)$. 
There is a natural constraint on the zero-order origin distance of the pdf
$\int_{-\infty}^{+\infty}dx p(x) = 1$
based on the normalization condition of the pdf. 
In addition, the contraints on the mean and variance yield,
$\int_{-\infty}^{+\infty} dx (xp(x))=\mu$, $\int_{-\infty}^{+\infty} dx (x^2p(x))=\sigma^2+\mu^2$, respectively. 
According to the variational principle, the above problem of conditional maximization
\begin{eqnarray*}
    \max_{p(x)} &&S[p]=-\int_{-\infty}^{+\infty} dx \left(p(x) \ln p(x)\right),\\ s.t.,~ &&\int_{-\infty}^{+\infty}dx p(x)= 1,~\int_{-\infty}^{+\infty} dx (xp(x))=\mu,~\int_{-\infty}^{+\infty} dx (x^2p(x))=\sigma^2+\mu^2,
\end{eqnarray*}
turns into a unconditional maximization problem by using the Lagrangian multiplier method,
\begin{equation}\label{H1}
\begin{aligned}
   \max_{p(x);\lambda_i}L[p;\lambda_i]
    =&S[p]+\lambda_0\left(\int_{-\infty}^{+\infty}dx p(x) - 1 \right) +\lambda_1\left(
    \int_{-\infty}^{+\infty} dx xp(x)-\mu\right)\\
    +&\lambda_2\left(\int_{-\infty}^{+\infty} dx x^2p(x)-\sigma^2-\mu^2\right),
\end{aligned}
\end{equation}
where $\lambda_i$ are Lagrangian multipliers for the corresponding constraints ($i=0,1,2$). Therefore, the variation (resp. partial differential) of $L$ with respect to $p(x)$ (resp. to  $\lambda_i$) vanishes, 
\begin{equation}
    0=\frac{\delta L}{\delta p}=\frac{\partial L}{\partial \lambda_i},~(i=0,1,2),
\end{equation}
which yields an optimal pdf as $p^*(x)=\exp(\lambda_0 -1 + \lambda_1 x + \lambda_2 x^2)$ with three algebraic relations 
\begin{equation*}
\begin{aligned}
&e^{\lambda_0 - 1} \int_{-\infty}^{+\infty}dx  \exp(\lambda_1 x + \lambda_2 x^2) =1, \\
&e^{\lambda_0 - 1} \int_{-\infty}^{+\infty}dx  x\exp(\lambda_1 x + \lambda_2 x^2) =\mu,\\
&e^{\lambda_0 - 1} \int_{-\infty}^{+\infty}dx  x^2 \exp(\lambda_1 x + \lambda_2 x^2)=\sigma^2+\mu^2.
\end{aligned}   
\end{equation*}
By utilizing the variable substitution and the integrals  $\int_{-\infty}^{+\infty} dx e^{-x^2} =\int_{0}^{+\infty} dx x^{-\frac{1}{2}}e^{-x}=  \sqrt{\pi}$, (the latter is also known as the Gamma function $\Gamma(\frac{1}{2})$), we obtain 
\begin{align*}
    & \int_{-\infty}^{+\infty}dx  \exp(\lambda_1 x + \lambda_2 x^2) =\sqrt{\pi}(-\lambda_2)^{-\frac{1}{2}}\exp\left(-\frac{{\lambda_1^2}}{4\lambda_2}\right), \\
    &\int_{-\infty}^{+\infty}dx  x\exp(\lambda_1 x + \lambda_2 x^2)=\frac{1}{2}\sqrt{\pi}\lambda_1(-\lambda_2)^{-\frac{3}{2}}\exp\left(-\frac{{\lambda_1^2}}{4\lambda_2}\right), \\
    &\int_{-\infty}^{+\infty}dx  x^2 \exp(\lambda_1 x + \lambda_2 x^2) =\frac{1}{4}\sqrt{\pi}(\lambda_1^2 - 2\lambda_2)(-\lambda_2)^{-\frac{5}{2}}\exp\left(-\frac{{\lambda_1^2}}{4\lambda_2}\right),
\end{align*}
for $\lambda_2 < 0$. 
Then the Lagrangian multipliers are readily expressed in terms of the observables as
\begin{equation*}
\begin{split}
\lambda_2&
=-\frac{1}{2\sigma^2},\quad
\lambda_1
=\frac{\mu}{\sigma^2},\quad
\lambda_0
=-\frac{1}{2}\left[ (\frac{\mu}{\sigma})^2 + \ln \sigma^2 +\ln(2\pi) \right].
\end{split}
\end{equation*} 
As a result, the maximization of the BG entropy with measured mean and variance deduces the general normal distribution as 
\begin{equation}\label{normal}
p^*(x)=\frac{1}{\sqrt{2\pi}\sigma}\exp\left({-\frac{(x-\mu)^2}{2\sigma^2}}\right)\sim N(\mu, \sigma^2).
\end{equation}
In the following, the superscript $^*$ will be omitted for notation simplicity without misunderstanding. The maximum value of BG entropy becomes 
$$
S[p^*]=\int_{-\infty}^{+\infty}dx  \left[\frac{1}{\sqrt{2\pi}\sigma}\exp\left({-\frac{(x-\mu)^2}{2\sigma^2}}\right) \right]
\left[\ln(\sqrt{2\pi}\sigma)+\ln\left({\frac{(x-\mu)^2}{2\sigma^2}}\right) \right]
=\ln(\sqrt{2\pi e}\sigma),
$$ 
here $e$ denotes the natural constant. 
The necessary condition can be verified by reversing the above derivation process. This completes the proof. 
\end{proof}

\begin{rem}
The pdf of normal distributions in Eq. \eqref{normal} can be rewritten into observables as $p^*(x)=\frac{1}{\sqrt{2\pi(\langle X^2 \rangle - \langle X \rangle^2)}}\exp\left({-\frac{(x - \langle X \rangle)^2}{2(\langle X^2 \rangle - \langle X \rangle^2)}}\right)$.
\end{rem}

\begin{rem}
Notice that the constants $1, \mu, \sigma^2$ have no effect on the variational principle w.r.t. $p(x)$. The augmented objective function in Eq. \eqref{H1} in some references is simplified as 
\begin{equation*}
\begin{aligned}
   \max_{p(X)}H[p]
    =&S[p]+\lambda_0\int_{-\infty}^{+\infty}dx p(x) +\lambda_1
    \int_{-\infty}^{+\infty} dx (xp(x))
    +\lambda_2\int_{-\infty}^{+\infty} dx (x^2p(x)).
\end{aligned}
\end{equation*}
However, in this formulation one has to impose constraints on parameters $\lambda_i$ separately. 
\end{rem}

\begin{rem}
\emph{(Generalization to multi-dimensions)} 
For a $n$-d random variable $\vec{X} \in \mathbf{R}^n$ 
with two observables, the mean $\langle \vec{X} \rangle\equiv \vec{\mu}$ and the co-variance matrix $\langle (\vec{X}  - \langle \vec{X} \rangle) (\vec{X}  - \langle \vec{X} \rangle)^T \rangle \equiv \boldsymbol{\Sigma}$. 
Then if and only if the probability density function is normally distributed with its mean equaling to $\vec{\mu}$ and its co-variance equaling to $\boldsymbol{\Sigma}$, i.e., $p^*(\vec{x})=\frac{1}{(2\pi)^{\frac{n}{2}}|\boldsymbol{\Sigma}|^{\frac{1}{2}}}\exp\left({-\frac{1}{2}(\vec{x}-\vec{\mu})^T {\boldsymbol{\Sigma}}^{-1} (\vec{x}-\vec{\mu}) }\right)$, the BG entropy function reaches its maximum value  $S[p^*]=-\int_{-\infty}^{+\infty} d\vec{x} p^*(\vec{x}) \ln p^*(\vec{x})=\ln\left[(2\pi e)^{\frac{n}{2}}{|\boldsymbol{\Sigma}|}^{\frac{1}{2}}\right]$.
\end{rem}

For the \textbf{exponential} distribution, assume that the pdf is zero when $x\leq 0$, then the augmented objective function becomes $L[p;\lambda_i] = -\int_{0}^{\infty} dx(p \ln p)+\sum_{i=0}^{1}\lambda_i\left(\int_{0}^{\infty} dx p x^i - \langle X^i \rangle \right)$. Standard calculations show that the optimal distribution is exponential, as
$p^*(x)=\frac{1}{\langle X \rangle}\exp\left(-\frac{x}{\langle X \rangle}\right)$, ${x>0}$.

For \textbf{higher-order} distribution, i.e., the maximisation of the BG entropy with constraints on the original distance up to order $M\geq 3$, we have 
$$ 
\max_{p(x);\lambda_i}L[p;\lambda_i]
    =-\int_{-\infty}^{+\infty} dx(p \ln p)
+\sum_{i=0}^{M}\lambda_i\left(\int_{-\infty}^{+\infty} dx p x^i - \langle X^i \rangle \right),
$$
which results in an optimal distribution as $p^*(x)=\exp(\lambda_0 - 1)\exp\left(\sum_{i=1}^{M} \lambda_i x^i \right)$. Here the $M$ parameters $\lambda_1, \cdots, \lambda_M$ are determined by $\exp(\lambda_0 - 1)\int_{-\infty}^{+\infty} dx \exp\left(\sum_{i=1}^{M} \lambda_i x^i \right) x^i = \langle X^i \rangle$, while $\lambda_0$ is determined by the normalization condition. 

For the \textbf{Gamma} distribution, the augmented objective function is chosen as $-\int_{0}^{\infty} dx(p \ln p)
+\sum_{i=0}^{1}\lambda_i\left(\int dx p x^i - \langle X^i \rangle \right)+\lambda_2\left(\int dx p \ln x - \langle \ln X \rangle \right)$, the optimal distribution is $p^*(x)=e^{\lambda_0 -1}e^{\lambda_1 x}x^{\lambda_2},~x\geq 0$. Using the integration
\begin{align*}
    & e^{\lambda_0 -1}\int_{0}^{\infty} dx e^{\lambda_1 x}x^{\lambda_2}= e^{\lambda_0 -1}(-\lambda_1)^{-\lambda_2-1} \Gamma(\lambda_2+1)=1, \\
    & e^{\lambda_0 -1}\int_{0}^{\infty} dx e^{\lambda_1 x}x^{\lambda_2+1}=e^{\lambda_0 -1} (-\lambda_1)^{-\lambda_2-2} \Gamma(\lambda_2+2)=
\langle X \rangle, \\
    &e^{\lambda_0 -1}\int_{0}^{\infty} dx e^{\lambda_1 x}x^{\lambda_2} \ln x=e^{\lambda_0 -1}(-\lambda_1)^{-\lambda_2-1} \left(\Gamma^{'}{(\lambda_2+1)} - \Gamma(\lambda_2+1) \ln(-\lambda_1) \right)=\langle \ln X \rangle,
\end{align*}
for $\lambda_1 <0, \lambda_2>-1$, here $\Gamma(\cdot)$ denotes the Gamma function,  $\Gamma^{'}(\cdot)$ denotes the derivative. 
The relation can be further obtained from $e^{\lambda_0-1}=(-\lambda_1)^{\lambda_2+1} /\Gamma(\lambda_2+1)$, $\lambda_1=-(\lambda_2 + 1)/\langle X \rangle$, $\ln(\lambda_2+1) - \Gamma^{'}{(\lambda_2+1)}/ \Gamma(\lambda_2+1)=\ln \langle X\rangle - \langle \ln X \rangle$. Since $\langle X\rangle$ and $ \langle \ln X \rangle$ are already known, we can derive the parameters $\lambda_i$ from these expressions. 
The optimal distribution is $p^*(x)=x^{\lambda_2}e^{\lambda_1 x} (-\lambda_1)^{\lambda_2+1}/\Gamma(\lambda_2+1),~x>0$.

Since the \textbf{Erlang} distribution is a particular case of the Gamma distribution with an integer shape parameter, i.e., $\lambda_2 \in \mathbb{Z}_{\geq 0}$, we can take the (approximate) integer solution of the equation, $\ln(\lambda_2+1) - \Gamma^{'}{(\lambda_2+1)}/ \Gamma(\lambda_2+1)=\ln \langle X\rangle - \langle \ln X \rangle$. The rest procedures are the same as that of the Gamma distribution. 

For the \textbf{Beta} distribution, the augmented objective function is chosen as  $-\int_{0}^{1} dx(p \ln p)+\lambda_0\left(\int dx p - 1\right)+\lambda_1\left(\int dx p \ln x - \langle \ln X \rangle \right)+\lambda_2\left(\int dx p \ln (1-x) - \langle \ln (1-X) \rangle \right)$, the optimal distribution is $e^{\lambda_0-1}x^{\lambda_1} (1-x)^{\lambda_2},~0<x<1$. Using the integration 
\begin{align*}
    & \int_{0}^{1} dx x^{\lambda_1} (1-x)^{\lambda_2}= \frac{\Gamma(\lambda_1+1)\Gamma(\lambda_2+1)}{\Gamma(\lambda_1+\lambda_2+2)}=e^{1-\lambda_0}, \\
   & \int_{0}^{1} dx x^{\lambda_1} (1-x)^{\lambda_2}\ln x= \frac{\Gamma(\lambda_1+1)\Gamma(\lambda_2+1)}{\Gamma(\lambda_1+\lambda_2+2)}\left(\frac{\Gamma^{'}(\lambda_1+1)}{\Gamma(\lambda_1+1)}-\frac{\Gamma^{'}(\lambda_1+\lambda_2+2)}{\Gamma(\lambda_1+\lambda_2+2)}\right)=e^{1-\lambda_0}
\langle \ln X \rangle, \\
    &\int_{0}^{1} dx x^{\lambda_1} (1-x)^{\lambda_2}\ln (1-x)= \frac{\Gamma(\lambda_1+1)\Gamma(\lambda_2+1)}{\Gamma(\lambda_1+\lambda_2+2)}\left(\frac{\Gamma^{'}(\lambda_2+1)}{\Gamma(\lambda_2+1)}-\frac{\Gamma^{'}(\lambda_1+\lambda_2+2)}{\Gamma(\lambda_1+\lambda_2+2)}\right)\\
    &=e^{1-\lambda_0}\langle \ln(1-X) \rangle,
\end{align*}
for $\lambda_1 >-1, \lambda_2>-1$. The optimal distribution becomes  $p^*(x)=\frac{\Gamma(\lambda_1+\lambda_2+2)}{\Gamma(\lambda_1+1)\Gamma(\lambda_2+1)}x^{\lambda_1}(1-x)^{\lambda_2},~0<x<1$, where the parameters $\lambda_1$ and $\lambda_2$ can be obtained from relations   $\frac{\Gamma^{'}(\lambda_1+1)}{\Gamma(\lambda_1+1)}-\frac{\Gamma^{'}(\lambda_1+\lambda_2+2)}{\Gamma(\lambda_1+\lambda_2+2)}=\langle \ln X \rangle,  \frac{\Gamma^{'}(\lambda_2+1)}{\Gamma(\lambda_2+1)}-\frac{\Gamma^{'}(\lambda_1+\lambda_2+2)}{\Gamma(\lambda_1+\lambda_2+2)}=\langle \ln(1-X) \rangle$. 

For the \textbf{Log-normal} distribution, 
$L[p;\lambda_i]=-\int_{0}^{+\infty} dx(p \ln p)+\lambda_0\left(\int dx p - 1\right)+\lambda_1\left(\int dx p \ln x - \langle \ln X \rangle \right)+\lambda_2\left(\int dx p (\ln x)^2 - \langle (\ln X)^2 \rangle \right)$, the resulting optimal pdf becomes  $p^*(x)=e^{\lambda_0-1}x^{\lambda_1} \exp\left[\lambda_2 (\ln x)^2\right]$ for $x>0$. Using the integration
\begin{align*}
    & \int_{0}^{+\infty} dx x^{\lambda_1} e^{[\lambda_2 (\ln x)^2]}= \sqrt{-\frac{\pi}{\lambda_2}}e^{-\frac{(\lambda_1+1)^2}{4\lambda_2}}=e^{1-\lambda_0}, \\
    & \int_{0}^{+\infty} dx x^{\lambda_1} \ln x e^{[\lambda_2 (\ln x)^2]}
    = \sqrt{-\frac{\pi}{\lambda_2}}e^{-\frac{(\lambda_1+1)^2}{4\lambda_2}} \left(\frac{\lambda_1+1}{2\lambda_2} \right)
    =e^{1-\lambda_0}
\langle \ln X \rangle, \\
    & \int_{0}^{+\infty} dx x^{\lambda_1} (\ln x)^2 e^{[\lambda_2 (\ln x)^2]}
    = \sqrt{-\frac{\pi}{\lambda_2}}e^{-\frac{(\lambda_1+1)^2}{4\lambda_2}} \left(\frac{(\lambda_1+1)^2}{4\lambda_2^2} - \frac{1}{2\lambda_2} \right)
    =e^{1-\lambda_0}
\langle (\ln X)^2 \rangle,
\end{align*}
for $\lambda_2<0$, the parameters are explicitly derived as 
\begin{equation*}
\begin{split}
\lambda_2&
=-\frac{1}{2}\frac{1}{\langle (\ln X)^2 \rangle - \langle\ln X \rangle^2},\quad
\lambda_1
=-1-2\lambda_2 \langle\ln X \rangle,\quad
\lambda_0
=1-\frac{1}{2}\ln{\left(-\frac{\pi}{\lambda_2}\right)} + \frac{(\lambda_1+1)^2}{4\lambda_2}.
\end{split}
\end{equation*} 
Thus, the optimal distribution becomes $p^*(x)=
\frac{1}{x \sqrt{2\pi} \sqrt{\langle (\ln X)^2 \rangle - \langle\ln X \rangle^2}}\exp{\left( \frac{(\ln x - \langle\ln X \rangle)^2}{-2(\langle (\ln X)^2 \rangle - \langle\ln X \rangle^2)}\right)}
,~0<x<\infty$. 

For the \textbf{double exponential} distribution, also called the Laplace distribution, whose symmetry axis is known as $x=\mu$, the Lagrangian is $L[p;\lambda_i]=-\int_{-\infty}^{+\infty} dx (p \ln p)+\lambda_0\left(\int dx p - 1 \right)+ \lambda_1\left(\int dx p|x-\mu| - \langle|X-\mu| \rangle \right)$. The resulting optimal pdf is $p^*(x)=e^{\lambda_0-1}e^{\lambda_1|x-\mu|}$ for $\lambda_1<0$. By itergration, we get $\lambda_1=-\frac{1}{\langle|X-\mu| \rangle}$, then the optiaml distribution becomes   $p^*(x)=\frac{1}{2\langle|X-\mu| \rangle}\exp{\left(-\frac{|x-\mu|}{\langle|X-\mu| \rangle} \right)}$.  

For the \textbf{Chi-squared} distribution, 
$L[p;\lambda_i]=-\int_{0}^{\infty} dx(p \ln p)+ \lambda_0\left(\int dx p - 1 \right)+\lambda_1\left(\int dx p x - \langle X \rangle \right)+\lambda_2\left(\int dx p \ln x - \frac{\Gamma'({\langle X \rangle}/2)}{\Gamma({\langle X \rangle}/2)} - \ln 2 \right)$. It is remarkable that the constrains of $\lambda_1$ and $\lambda_2$ share the same constant $\langle X \rangle$ via measurements. Then 
$p^*(x)=\frac{1}{2^{{\langle X \rangle}/{2}} \Gamma({\langle X \rangle}/{2})}x^{{\langle X \rangle}/{2}-1}e^{-{\langle X \rangle}/{2}} \mathbb{I}_{x>0}$.

For the \textbf{F}-distribution, 
\begin{align*}
L[p;\lambda_i]=
&-\int_{0}^{\infty} dx(p \ln p)+ \lambda_0\left(\int dx p - 1 \right)+\lambda_1\left(\int dx p \ln x - \langle \ln X \rangle \right)\\
&+\lambda_2\left(\int dx p \ln (1+\frac{d_1}{d_2}x) - \langle \ln (1+\frac{d_1}{d_2}X) \rangle \right),
\end{align*}
where $d_1$ and $d_2$ are two undetermined positive parameters. The resulting optimal pdf is 
$p^*(x)=e^{\lambda_0 -1} x^{\lambda_1} (1+\frac{d_1}{d_2}x)^{\lambda_2} \mathbb{I}_{x\geq 0}$, where the Lagrangian multipliers are obtained by solving 
\begin{align*}
    & \int_{0}^{\infty} dx x^{\lambda_1} (1+\frac{d_1}{d_2}x)^{\lambda_2}
    =e^{1-\lambda_0}, \\
   & \int_{0}^{\infty} dx x^{\lambda_1} (1+\frac{d_1}{d_2}x)^{\lambda_2}\ln x
    =e^{1-\lambda_0}\langle \ln X \rangle, \\
 & \int_{0}^{\infty} dx x^{\lambda_1} (1+\frac{d_1}{d_2}x)^{\lambda_2} \ln (1+\frac{d_1}{d_2}x)
 =e^{1-\lambda_0}\langle \ln (1+\frac{d_1}{d_2}X) \rangle.
\end{align*}

For the \textbf{Rayleigh} distribution, $L[p;\lambda_i]=-\int_{0}^{\infty} dx(p \ln p)+ \lambda_0\left(\int dx p - 1 \right)+\lambda_1\left(\int dx p x^2 - \langle X^2 \rangle \right)+\lambda_2\left(\int dx p \ln x - \langle \ln X \rangle \right)$. Then the optimal distribution is $p^*(x)=e^{\lambda_0 -1}e^{\lambda_1 x^2}x^{\lambda_2},~x\geq 0$. Using the integration 
\begin{align*}
    & \int_{0}^{\infty} dx e^{\lambda_1 x^2}x^{\lambda_2}=\frac{1}{2} (-\lambda_1)^{-\frac{\lambda_2+1}{2}} \Gamma(\frac{\lambda_2+1}{2})
    =e^{1-\lambda_0}, \\
    & \int_{0}^{\infty} dx e^{\lambda_1 x^2}x^{\lambda_2+2}=
    \frac{1}{2} (-\lambda_1)^{-\frac{\lambda_2+3}{2}} \Gamma(\frac{\lambda_2+3}{2})
    =e^{1-\lambda_0}\langle X^2 \rangle, \\
    &\int_{0}^{\infty} dx e^{\lambda_1 x^2}x^{\lambda_2} \ln x
    = \frac{1}{4}(-\lambda_1)^{-\frac{\lambda_2+1}{2}} \left(\Gamma^{'}{(\frac{\lambda_2+1}{2})} - \Gamma(\frac{\lambda_2+1}{2}) \ln(-\lambda_1) \right)=e^{1-\lambda_0}\langle \ln X \rangle,
\end{align*}
for $\lambda_1 <0, \lambda_2>-1$. 
The relation can be further obtained $\frac{1}{2} (-\lambda_1)^{-\frac{\lambda_2+1}{2}} \Gamma(\frac{\lambda_2+1}{2})=e^{1-\lambda_0}$, $\lambda_1=-(\lambda_2 + 1)/(2\langle X^2 \rangle)$, $\ln\left(\frac{\lambda_2+1}{2}\right) - \frac{1}{2}\Gamma^{'}{(\frac{\lambda_2+1}{2})}/ \Gamma(\frac{\lambda_2+1}{2})=\ln \langle X^2\rangle - \langle \ln X \rangle$. Since $\langle X^2\rangle$ and $ \langle \ln X \rangle$ are already known from measurements, we can derive the parameter $\lambda_2$ from the last expression, $\lambda_1$ from the middle one, and $\lambda_0$ from the first one. 

On the other hand, the standard expression of the Rayleigh distribution reads  $p(x)=\frac{x}{\sigma^2}exp(-\frac{x^2}{2\sigma^2})$ for $x\geq 0$, which means that $\lambda_2 = 1$ in $p^*(x)=e^{\lambda_0 -1}e^{\lambda_1 x^2}x^{\lambda_2}$. Thus, the difference between the solution $\lambda_2$ to the equality 
\begin{equation}
\label{Rayleigh}
\ln\left(\frac{\lambda_2+1}{2}\right) - \frac{1}{2}\Gamma^{'}{(\frac{\lambda_2+1}{2})}/ \Gamma(\frac{\lambda_2+1}{2})=\ln \langle X^2\rangle - \langle \ln X \rangle,
\end{equation}
and $1$, $d(\lambda_2,1)=|\lambda_2 - 1|$, is adopted as a test criteria whether the data can be fitted to the Rayleigh distribution. Actually, when we restrict that $\lambda_2=1$,  Eq. \eqref{Rayleigh} reduces to $\ln \langle X^2\rangle - \langle \ln X \rangle=-\frac{1}{2}\Gamma^{'}{(1)}$, where $-\Gamma^{'}{(1)}$ on the right-hand side is also known as the Euler–Mascheroni constant. 


For the \textbf{Weibull} distribution, 
$L[p;\lambda_i]=-\int_{0}^{\infty} dx(p \ln p)+ \lambda_0\left(\int dx p - 1 \right)+\lambda_1\left(\int dx p x^k - \langle X^k \rangle \right)+\lambda_2\left(\int dx p \ln x - \langle \ln X \rangle \right)$. 
The optimal distribution is $p^*(x)=e^{\lambda_0 -1}e^{\lambda_1 x^k}x^{\lambda_2},~x\geq 0$. Using the integration 
\begin{align*}
    & \int_{0}^{\infty} dx e^{\lambda_1 x^k}x^{\lambda_2}=\frac{1}{k} (-\lambda_1)^{-\frac{\lambda_2+1}{k}} \Gamma(\frac{\lambda_2+1}{k})
    =e^{1-\lambda_0}, \\
    & \int_{0}^{\infty} dx e^{\lambda_1 x^k}x^{\lambda_2+k}=
    \frac{1}{k} (-\lambda_1)^{-\frac{\lambda_2+k+1}{k}} \Gamma(\frac{\lambda_2+k+1}{k})
    =e^{1-\lambda_0}\langle X^k \rangle, \\
    &\int_{0}^{\infty} dx e^{\lambda_1 x^k}x^{\lambda_2} \ln x
    = \frac{1}{k^2}(-\lambda_1)^{-\frac{\lambda_2+1}{k}} \left(\Gamma^{'}{(\frac{\lambda_2+1}{k})} - \Gamma(\frac{\lambda_2+1}{k}) \ln(-\lambda_1) \right)=e^{1-\lambda_0}\langle \ln X \rangle,
\end{align*}
for $\lambda_1 <0, \lambda_2>-k$,  
the relation can be further obtained $\frac{1}{k} (-\lambda_1)^{-\frac{\lambda_2+1}{k}} \Gamma(\frac{\lambda_2+1}{k})=e^{1-\lambda_0}$, $\lambda_1=-(\lambda_2 + 1)/(k\langle X^k \rangle)$,  $\ln\left(\frac{\lambda_2+1}{k}\right) - \Gamma^{'}{(\frac{\lambda_2+1}{k})}/ \Gamma(\frac{\lambda_2+1}{k})=\ln \langle X^k\rangle - k\langle \ln X \rangle$. 
The standard Weibull distribution $p(x)=\frac{k}{\lambda^k} x^{k-1} \exp(-\frac{x^k}{\lambda^k})$ requires that $\lambda_2=k-1$, with $k>0, \lambda>0$ representing the shape parameter and scale parameter respectively.  

For the \textbf{generalized error} distribution (GED), also called as the symmetric generalized normal distribution, we have
\begin{align*}
L[p;\lambda_i]=
&-\int_{-\infty}^{\infty} dx(p \ln p)+ \lambda_0\left(\int dx p - 1 \right)+\lambda_1\left(\int dx p |x-\mu|^{\beta} - \langle  |X-\mu|^{\beta} \rangle \right),
\end{align*}
where $\mu$ and $\beta$ are two undetermined parameters $(\beta>0)$. The resulting optimal pdf is $p^*(x)=e^{\lambda_0 -1} e^{\lambda_1|x-\mu|^{\beta}}$, where the Lagrangian multipliers are obtained by solving 
\begin{align*}
    & \int_{-\infty}^{\infty} dx e^{\lambda_1|x-\mu|^{\beta}}
    =e^{1-\lambda_0}, \\
   & \int_{-\infty}^{\infty} dx e^{\lambda_1|x-\mu|^{\beta}} |x-\mu|^{\beta}
    =e^{1-\lambda_0} \langle |X-\mu|^{\beta} \rangle.
\end{align*}

For the standard \textbf{Cauchy} distribution, 
\begin{align*}
L[p;\lambda_i]=
&-\int_{-\infty}^{\infty} dx(p \ln p)+ \lambda_0\left(\int dx p - 1 \right)+\lambda_1\left(\int dx p \ln (1+x^2) - \langle \ln (1+X^2) \rangle \right).
\end{align*}
The resulting optimal pdf is 
$p^*(x)=e^{\lambda_0 -1} (1+x^2)^{\lambda_1}$, where the Lagrangian multipliers are obtained by solving 
\begin{align*}
    & \int_{-\infty}^{\infty} dx (1+x^2)^{\lambda_1}
    =e^{1-\lambda_0}, \\
   & \int_{-\infty}^{\infty} dx (1+x^2)^{\lambda_1} \ln (1+x^2)
    =e^{1-\lambda_0} \langle \ln (1+X^2) \rangle.
\end{align*}

For the \textbf{Student's t} distribution, 
\begin{align*}
L[p;\lambda_i]=
&-\int_{-\infty}^{\infty} dx(p \ln p)+ \lambda_0\left(\int dx p - 1 \right)+\lambda_1\left(\int dx p \ln (1+{\nu}^{-1}x^2) - \langle \ln (1+{\nu}^{-1}X^2) \rangle \right),
\end{align*}
where $\nu>0$ is a undetermined parameter. The resulting optimal pdf is 
$p^*(x)=e^{\lambda_0 -1} (1+{\nu}^{-1}x^2)^{\lambda_1}$, where the Lagrangian multipliers are obtained by solving 
\begin{align*}
    & \int_{-\infty}^{\infty} dx (1+{\nu}^{-1}x^2)^{\lambda_1}
    =e^{1-\lambda_0}, \\
   & \int_{-\infty}^{\infty} dx (1+{\nu}^{-1}x^2)^{\lambda_1} \ln (1+{\nu}^{-1}x^2)
    =e^{1-\lambda_0} \langle \ln (1+{\nu}^{-1}X^2) \rangle.
\end{align*} 
Notice that the Student's t distribution reduces to the standard Cauchy distribution when $\nu=1$. 

For the \textbf{Pearson type-IV} distribution, 
\begin{align*}
L[p;\lambda_i]=
&-\int_{-\infty}^{\infty} dx(p \ln p)+ \lambda_0\left(\int dx p - 1 \right)+\lambda_1\left(\int dx p \ln ( 1+(\frac{x-\lambda}{\alpha})^2 ) - \langle \ln ( 1+(\frac{X-\lambda}{\alpha})^2 ) \rangle \right)\\
&+\lambda_2\left(\int dx p \arctan(\frac{x-\lambda}{\alpha}) - \langle \arctan(\frac{X-\lambda}{\alpha}) \rangle \right),
\end{align*}
where $\lambda$ and $\alpha$ are two undetermined positive parameters. The resulting optimal pdf is 
$p^*(x)=e^{\lambda_0 -1} [ 1+(\frac{x-\lambda}{\alpha})^2 ]^{\lambda_1} \exp[\lambda_2 \arctan(\frac{x-\lambda}{\alpha})]$, where the Lagrangian multipliers are obtained by solving 
\begin{align*}
    & \int_{-\infty}^{\infty} dx [ 1+(\frac{x-\lambda}{\alpha})^2 ]^{\lambda_1} \exp[\lambda_2 \arctan(\frac{x-\lambda}{\alpha})]
    =e^{1-\lambda_0}, \\
   & \int_{-\infty}^{\infty} dx [ 1+(\frac{x-\lambda}{\alpha})^2 ]^{\lambda_1} \exp[\lambda_2 \arctan(\frac{x-\lambda}{\alpha})] \ln [ 1+(\frac{x-\lambda}{\alpha})^2 ]
    =e^{1-\lambda_0}\langle \ln [ 1+(\frac{X-\lambda}{\alpha})^2 ] \rangle, \\
 & \int_{-\infty}^{\infty} dx [ 1+(\frac{x-\lambda}{\alpha})^2 ]^{\lambda_1} \exp[\lambda_2 \arctan(\frac{x-\lambda}{\alpha})] \arctan(\frac{x-\lambda}{\alpha})
 =e^{1-\lambda_0}\langle \arctan(\frac{X-\lambda}{\alpha}) \rangle.
\end{align*}

For the \textbf{generalized Student's t} distribution \cite{park2009maximum}, 
\begin{align*}
L[p;\lambda_i]=
&-\int_{-\infty}^{\infty} dx(p \ln p)+ \lambda_0\left(\int dx p - 1 \right)+\lambda_1\left(\int dx p \ln (1+{\nu}^{-1}x^2) - \langle \ln (1+{\nu}^{-1}X^2) \rangle \right)\\
&+\lambda_2\left(\int dx p \arctan(\frac{x}{\alpha}) - \langle \arctan(\frac{X}{\alpha}) \rangle \right) 
+\sum_{i=3}^{M}\lambda_{i}\left(\int_{-\infty}^{+\infty} dx p x^{i-2} - \langle X^{i-2} \rangle \right),
\end{align*}
where $\nu$ and $\alpha$ are two undetermined positive parameters. The resulting optimal pdf is $p^*(x)=e^{\lambda_0 -1} (1+{\nu}^{-1}x^2)^{\lambda_1} \exp[\lambda_2 \arctan(\frac{x}{\alpha})] \exp\left(\sum_{i=3}^{M} \lambda_i x^{i-2} \right)$, where the Lagrangian multipliers are obtained by solving 
\begin{align*}
    & \int_{-\infty}^{\infty} dx (1+{\nu}^{-1}x^2)^{\lambda_1} \exp[\lambda_2 \arctan(\frac{x}{\alpha})] \exp\left(\sum_{i=3}^{M} \lambda_i x^{i-2} \right)
    =e^{1-\lambda_0}, \\
   & \int_{-\infty}^{\infty} dx (1+{\nu}^{-1}x^2)^{\lambda_1} \exp[\lambda_2 \arctan(\frac{x}{\alpha})] \exp\left(\sum_{i=3}^{M} \lambda_i x^{i-2} \right) \ln (1+{\nu}^{-1}x^2)
    =e^{1-\lambda_0} \langle \ln (1+{\nu}^{-1}X^2) \rangle, \\
   & \int_{-\infty}^{\infty} dx (1+{\nu}^{-1}x^2)^{\lambda_1} \exp[\lambda_2 \arctan(\frac{x}{\alpha})] \exp\left(\sum_{i=3}^{M} \lambda_i x^{i-2} \right) arctan(\frac{x}{\alpha}) 
    =e^{1-\lambda_0} \langle \arctan(\frac{X}{\alpha}) \rangle, \\
   & \int_{-\infty}^{\infty} dx (1+{\nu}^{-1}x^2)^{\lambda_1} \exp[\lambda_2 \arctan(\frac{x}{\alpha})] \exp\left(\sum_{i=3}^{M} \lambda_i x^{i-2} \right) x^{i-2}
    =e^{1-\lambda_0} \langle X^{i-2} \rangle, \quad i=3,\cdots, M.
\end{align*} 

For the \textbf{generalized Log-normal} distribution  \cite{park2009maximum}, 
\begin{align*}
L[p;\lambda_i]=
&-\int_{0}^{+\infty} dx(p \ln p)+\lambda_0\left(\int dx p - 1\right)+\lambda_1\left(\int dx p \ln x - \langle \ln X \rangle \right)\\
&+\lambda_2\left(\int dx p (\ln x)^2 - \langle (\ln X)^2 \rangle \right) + \sum_{i=3}^{M}\lambda_{i}\left(\int_{-\infty}^{+\infty} dx p x^{i-2} - \langle X^{i-2} \rangle \right).
\end{align*} 
The resulting optimal pdf becomes  $p^*(x)=e^{\lambda_0-1}x^{\lambda_1} \exp\left[\lambda_2 (\ln x)^2\right] \exp\left(\sum_{i=3}^{M} \lambda_i x^{i-2} \right)$ for $x>0$, where the Lagrangian multipliers are obtained by solving 
\begin{align*}
    & \int_{0}^{+\infty} dx x^{\lambda_1} e^{[\lambda_2 (\ln x)^2]} \exp\left(\sum_{i=3}^{M} \lambda_i x^{i-2} \right)=e^{1-\lambda_0}, \\
    & \int_{0}^{+\infty} dx x^{\lambda_1} \ln x e^{[\lambda_2 (\ln x)^2]} \exp\left(\sum_{i=3}^{M} \lambda_i x^{i-2} \right)
    =e^{1-\lambda_0}
\langle \ln X \rangle, \\
    & \int_{0}^{+\infty} dx x^{\lambda_1} (\ln x)^2 e^{[\lambda_2 (\ln x)^2]} \exp\left(\sum_{i=3}^{M} \lambda_i x^{i-2} \right)
    =e^{1-\lambda_0}\langle (\ln X)^2 \rangle,\\
    & \int_{0}^{+\infty} dx x^{\lambda_1+i-2} e^{[\lambda_2 (\ln x)^2]} \exp\left(\sum_{i=3}^{M} \lambda_i x^{i-2} \right)
    =e^{1-\lambda_0} \langle X^{i-2} \rangle, \quad i=3,\cdots, M.
\end{align*}

For the \textbf{L\'{e}vy} Distribution, Tsallis et al. \cite{tsallis1995statistical} generalized the BG entropy to a non-extensive one as 
\begin{equation}
S^T_q(t)
=\frac{1}{q-1}\left(1-\int d(\frac{x}{\sigma})(\sigma p)^q\right),
\end{equation} 
where the superscript of $S^T$ denotes the Tsallis entropy, $q \in \mathbb{R}$ denotes the degree of nonextensivity, and  $\sigma>0$ denotes the characteristic length. In addition to the normalization constraint $\int dx p = 1$, there is also a constraint on the $q$-expectation value of $x^2$,  
$\int d(\frac{x}{\sigma}) (\sigma p)^{q} x^2 = \sigma^2$. The resulting distribution becomes 
$\frac{1}{\sigma} \left(\frac{q}{\lambda_0 (1-q)} \right)^{\frac{1}{1-q}} \left(1-\lambda_1(1-q)x^2 \right)^{\frac{1}{1-q}}$, where $\frac{1}{\sigma} \left(\frac{q}{\lambda_0 (1-q)} \right)^{\frac{1}{1-q}}$ is the partition function. The MEP of the Tsallis entropy and the extension of central limit theorem provide an interpretation for the ubiquity of L\'{e}vy distribution \cite{tsallis1995statistical}. 

For the \textbf{stretched exponential} distribution, Anteneodo and Plastino \cite{anteneodo1999maximum} introduced a new non-extensive entropy  
\begin{equation}
S^{AR}_\eta(t)=
\int dx\left( \Gamma(\frac{\eta+1}{\eta}, -\ln p) -p\Gamma(\frac{\eta+1}{\eta}) \right),
\end{equation}
by incorporating the complementary incomplete Gamma function $\Gamma(\mu, t) = \int_0^{\exp(-t)} dx (-\ln x )^{\mu-1}$, where the superscript of $S^{AR}$ is used to distinguish this entropy from the others, $\eta>0$ is the stretching exponent. 
The constraints read  
$\lambda_0: \left(\int dx p - 1\right)$, 
$\lambda_i: \mathbb{E}({f}_{i}(x)) =  \langle {f}_{i}(X) \rangle$ for $i=1,\cdots, M$. The resulting distribution is 
$p^*(x)=\exp\left\{-\left[\lambda_0 + \sum_{i=1}^{M}\lambda_i {f}_{i}(x) + \Gamma(\frac{\eta+1}{\eta})\right]^{\eta} \right\}$. 

\section{Rigorous derivations of various discrete distribution functions by MEP}
\label{Rigorous derivations of various discrete distribution}
For the \textbf{discrete Maxwell-Boltzmann} distribution with $N$ microstates of energies, the BG entropy is 
$\sum_{i=1}^N \left(-p_i \ln p_i\right)$. The first  constraint is on the probability normalization, $\lambda_0\left(\sum_{i=1}^N p_i - 1 \right)$, and the second one is on the averaged energy $\lambda_1\left(\sum_{i=1}^N p_i x_i - \langle X \rangle \right)$. Combining these constraints with the entropy, we deduce the resulting probability as 
$p^*(x)=\exp( \lambda_1 x_i)/{\sum_{i=1}^N \exp( \lambda_1 x_i)}$, where the parameter $\lambda_1$ is determined by $\sum_{i=1}^N (x_i -\langle X \rangle)\exp(\lambda_1 x_i)=0$. 

Different from the distributions mentioned above, the following binomial and Poisson distributions are proved to be the maximum entropy distributions on some suitably defined sets \cite{harremoes2001binomial}. We present the results here and omit the proof. 
Denote $X_1, \cdots, X_N$ be a sequence of independent non-identical Bernoulli random variables with probabilities being $q_i=P(X_i=1)$. Denote the $N$-generalized binomial distribution $\sum_i^N X_i$ with the same mean $\lambda$ as $B_N(\lambda)$. Denote the union of all $N$-generalized binomial distributions with mean $\lambda$ ($N=1,2,\cdots$) as $B_{\infty}(\lambda) \equiv \cup_N B_N(\lambda)$. That is, 
\begin{align*}
B_N(\lambda)=&\left\{\sum_i^N X_i | \{X_i\}_{i=1}^N~\textit{independent},~\mathbb\sum_i^N q_i=\lambda \right\}, \\
B_{\infty}(\lambda) \equiv& \bigcup_{N=1}^{\infty} B_N(\lambda).
\end{align*}

For the \textbf{Binomial} distribution \cite{harremoes2001binomial}, the BG entropy is 
$\sum_{i=1}^N \left(-p_i \ln p_i\right)$. The binomial distribution $b(N, \frac{\lambda}{N})$ is the maximum-entropy distribution on the set $B_N(\lambda)$. 

For the \textbf{Poisson} distribution \cite{harremoes2001binomial}, the BG entropy is 
$\sum_{i=1}^{\infty} \left(-p_i \ln p_i\right)$. The Poisson distribution is the maximum-entropy distribution on the set $B_{\infty}(\lambda)$. 
Notice that the binomial distribution converges to the Poisson distribution when $N\rightarrow \infty$ and the mean of binomial distribution $Np$ converges to a finite limit. 

For the \textbf{geometric} distribution supported on $\{1,2,\cdots\}$, the BG entropy is 
$\sum_{i=1}^{\infty} \left(-p_i \ln p_i\right)$, and 
$L[p;\lambda_i]=\sum_{i=1}^{\infty} \left(-p_i \ln p_i\right) + \lambda_0\left(\sum_{i=1}^{\infty} p_i - 1 \right)+\lambda_1\left(\sum_{i=1}^{\infty} i p_i - \langle X \rangle \right)$. 
The optimal distribution is $p^*_i=e^{\lambda_0 -1}e^{\lambda_1 i},~i=1,2,\cdots$. 
Based on  
\begin{align*}
    &\sum_{i=1}^{\infty} e^{\lambda_1 i}
    =\frac{e^{\lambda_1}}{1- e^{\lambda_1}}
    =e^{1-\lambda_0}, \quad 
    \sum_{i=1}^{\infty} e^{\lambda_1 i} i  
    =\frac{e^{\lambda_1}}{(1- e^{\lambda_1})^2}
    =e^{1-\lambda_0}\langle X \rangle, 
\end{align*}
for $\lambda_1 <0$, the optimal distribution becomes $p^*_i=\frac{1}{\langle X \rangle} \left(1 - \frac{1}{\langle X \rangle}\right)^{i-1}$, which is the standard geometric distribution with $0<\frac{1}{\langle X \rangle}<1$, guaranteed by the relations of $\frac{1}{\langle X \rangle}=1-e^{\lambda_1}$ and $\lambda_1 <0$, being the success probability of a trail. 
\bibliography{aipsamp}%
\end{document}